\documentclass[journal,twocolumn,10pt]{IEEEtran}
\usepackage{framed,multicol,nomencl,etoolbox}
\usepackage{cuted}
\usepackage{amsfonts}
\usepackage{amsmath,cases,amssymb}
\usepackage[dvips]{graphicx}
\usepackage{cite,color}
\usepackage{subfigure}
\usepackage{color}
\usepackage{epsfig}
\usepackage{epstopdf}
\usepackage{url}
\usepackage{algorithm}
\usepackage{algorithmic}
\usepackage{color}
\usepackage{multirow}
\usepackage{mathtools}
\usepackage{hyperref}
\usepackage{soul}

\newcommand{\clL}{{\cal L}}

\newcommand{\ds}{\displaystyle}

\newcommand{\clD}{{\cal D}}

\newcommand{\clF}{{\cal F}}

\newcommand{\clC}{{\cal C}}

\newcommand{\clN}{{\cal N}}

\newcommand{\bgeqn}{\begin{equation}}
\newcommand{\edeqn}{\end{equation}}

\newcommand{\beqa}{\begin{eqnarray}}
\newcommand{\eeqa}{\end{eqnarray}}
\newcommand{\beqas}{\begin{eqnarray*}}
\newcommand{\eeqas}{\end{eqnarray*}}

\newcommand{\bm}{\mathbf{m}}

\newcommand{\br}{\mathbf{r}}
\newcommand{\bw}{\mathbf{w}}
\newcommand{\bv}{\mathbf{v}}
\makenomenclature

\renewcommand\nomgroup[1]{%
  \item[\bfseries
  \ifstrequal{#1}{A}{Abbreviations}{%
  \ifstrequal{#1}{S}{Sets}{%
  \ifstrequal{#1}{N}{Notations}}}%
]}

\begin{document}
\title{Hierarchical fuzzy neural networks with privacy preservation for heterogeneous big data}
\author{Leijie Zhang, Ye Shi, \textit{Member, IEEE}, Yu-Cheng Chang, and Chin-Teng Lin \textit{Fellow, IEEE}
\thanks{Leijie Zhang and Ye Shi contributed equally to this work. This work was supported in part by the Australian Research Council (ARC) under discovery grant DP180100670 and DP180100656. Research was also sponsored in part by the Australia Defence Innovation Hub under Contract No. P18-650825, and US Office of Naval Research Global under Cooperative Agreement Number ONRG-NICOP-N62909-19-1-2058. We also thank the NSW Defence Innovation Network and NSW State Government of Australia for financial support in part of this research through grant DINPP2019 S1-03/09. (Corresponding author: Ye Shi.)}
\thanks{Leijie Zhang, Ye Shi, Yu-Cheng Chang and Chin-Teng Lin are with CIBCI lab, Australian Artificial Intelligence Institute, the School of Computer Science, University of Technology, Sydney, NSW 2007, Australia. Email:
Leijie.Zhang@student.uts.edu.au, Ye.Shi-1@uts.edu.au, Yu-Cheng.Chang@student.uts.edu.au, Chin-Teng.Lin@uts.edu.au.}
}
\date{}
\maketitle
\begin{abstract}
Heterogeneous big data poses many challenges in machine learning. {\color{black} Its enormous scale, high dimensionality, and inherent uncertainty} make almost every aspect of machine learning difficult, from providing enough processing power to maintaining model accuracy to protecting privacy. {\color{black}However, perhaps the most imposing problem is that big data is often interspersed with sensitive personal data. Hence, we propose a privacy-preserving hierarchical fuzzy neural network (PP-HFNN) to address these technical challenges while also alleviating privacy concerns.} The network is trained with a two-stage optimization algorithm, and the parameters at low levels of the hierarchy are learned with a scheme based on the well-known alternating direction method of multipliers, which does not reveal local data to other agents. Coordination at high levels of the hierarchy {\color{black}is handled by} the alternating optimization method, which converges very quickly. The entire training procedure is scalable, fast and does not suffer from gradient vanishing problems like the methods based on back-propagation. Comprehensive simulations conducted on both regression and classification tasks demonstrate the effectiveness of the proposed model. Our code is available online\footnote{\url{https://github.com/leijiezhang/PP_HFNN}}.
\end{abstract}



\section{Introduction}
Big data and its benefits have become a ubiquitous part of research in many areas, including social networks, the Internet of Things, commerce, astronomy, biology, medicine, and more \cite{qiu2016survey, tran2019privacy}. However, with increasing regularity, the gains being made seem to come at the threat of violations to the privacy and security of our personal data \cite{mohassel2017secureml}. This issue is attracting serious attention  in  both  society and the research community, especially post the Facebook data privacy scandal \cite{tuttle2018facebook}. In addition to privacy concerns, heterogeneous big data and its inherent uncertainty, enormous scales and high dimensionality are posing even more challenges to traditional machine learning methods –-- for instance, the extraordinary amounts of computing power needed to process big data and the negative effects uncertainty can have on a model’s accuracy and performance \cite{lewis1994heterogeneous,abadi2016tensorflow}. We need novel and efficient machine learning techniques to cope with all these challenges.

Usually, data heterogeneity stems from the nature of the information to be captured and/or the methods used to generate or acquire the data, and it can exist at either the sample, {\color{black}the feature level or both}. Traditional data processing methods, such as data cleaning, data integration, dimension reduction and data normalization, may need to be applied in combination to ensure {\color{black} they effectively reduce} data disparity \cite{wang2017heterogeneous}. In addition to heterogeneity, uncertainty is another problematic characteristic of heterogeneous big data.  Loosely defined as “how well the data speaks to the question of interest with respect to the model”\cite{osti1413588}, uncertainty over whether the available data will meet the demands of the task begins at the collection stage. {\color{black} Neither sensors nor} humans are immune to measurement errors; qualitative data, such as social media accounts, are often subjective in nature; and the data collected may not be relevant or might be incomplete.
Currently, a powerful and effective solution for dealing with uncertainty is to build the model through a fuzzy neural network (FNN), which involves fuzzy rules and fuzzy inference systems \cite{jang1993anfis,zuo2016fuzzy}, to make optimal use of the given data \cite{fu2011fuzzy,couso2019fuzzy}. FNNs deal with data uncertainty through ‘fuzzification’ operations and architectures based on if-then rules. However, in a standard fuzzy system, the number of fuzzy rules increases exponentially with the number of system variables \cite{raju1991hierarchical,zeng1995approximation,wang1999analysis}. This limits the feasibility of applying standard FNNs to heterogeneous big data.
In an attempt to overcome the rule explosion problem, Raju, as far back as 1991, proposed {\color{black}the first hierarchical fuzzy system} \cite{raju1991hierarchical}. It consisted of a number of {\color{black}low-dimensional} fuzzy systems connected in a hierarchical structure such that the total number of rules increased linearly with the number of input variables. In the years since then, hierarchical fuzzy systems have been studied in depth \cite{wang1999analysis,campello2006hierarchical,chen2007automatic,wang2019fast} and applied to many practical problems, such as price negotiation \cite{fu2015fuzzy}, video de-interlacing \cite{brox2010tuning}, linguistic attribute hierarchy \cite{he2014linguistic}, weapon target assignment\cite{csahin2011hierarchical}, cloud resources demand prediction \cite{chen2019prediction}, {\color{black}and more}. However, there is very little literature on using hierarchical fuzzy neural network (HFNN) to address the {\color{black}uncertainty, privacy concerns and computational overhead } associated with big data.

As one of the most salient problems in data science, privacy-preserving machine learning is attracting attention \cite{agrawal2000privacy,chaudhuri2009privacy,xu2015privacy, papernot2018sok}. Generally, there are two types of privacy-preserving machine learning methods; one relies on perturbation and randomization, the other on segmenting the data. {\color{black}Perturbation/randomization} methods alter the data before releasing or sharing them \cite{chaudhuri2009privacy}, which offers some limited privacy protection at the sacrifice of some performance. Methods based on segmentation typically distribute the data among multiple agents. Neighboring agents cooperate with each other to learn global results but without revealing their individual data to others \cite{clifton2002tools}.  Distributed machine learning algorithms have emerged out of these multi-agent data sharing scenarios as a solution to preserving privacy  \cite{forero2010consensus,qin2016distributed,bi2015distributed,ye2019decentralized,scardapane2016decentralized}.
{\color{black} Distributed algorithms are well suited to big data environments given there are limits on the amount of data a single agent or a centralized algorithm can feasibly handle. To harness the communications overhead and storage capacity needed to process information at the big data scale, the data needs to be distributed throughout a network of interconnected agents\cite{georgopoulos2014distributed}.}

This work brings {\color{black} the advantages of both hierarchical and distributed architectures to an FNN to tackle the privacy, uncertainty and feasibility constraints} associated with heterogeneous big data. HFNNs are specifically designed to extract features from heterogeneous  data  and to reduce the  number of fuzzy rules without compromising performance, {\color{black} which helps to overcome the potential exponential rule explosion.}  The system variables of the heterogeneous data are divided into several independent subsets, each of which only consists of homogeneous data that is then associated with an FNN at the lower levels of the hierarchy. The outputs of the low-level FNNs are coordinated at the higher levels of the hierarchy. For the purposes of this paper, we selected the alternating optimization (AO) method for high-level coordination. This is a widely-used method that sequentially optimizes {\color{black}the} objective function over one variable while fixing the other \cite{wang2008new,netrapalli2013phase,lu2019pa}. Compared to gradient-based algorithms, AO has several merits:
(i) {\color{black} At each iteration, AO} divides a difficult problem into much easier sub-problems, which usually have a closed-form solution; (ii) {\color{black}AO} is easy to implement as there is no need to tune {\color{black}the} optimization parameters, like the learning rate {\color{black}or} step sizes;  and (iii) it converges very quickly in practice and does not suffer from the vanishing gradient problem {\color{black}associated with} gradient-based learning methods.
Further, each low-level FNN operates in a distributed computation framework to combat the challenges {\color{black}raised by numerous data samples as well as to alleviate privacy concerns}. More specifically, a distributed clustering method, built on the well-known alternating direction method of multipliers (ADMM), locates the parameters {\color{black}in} the antecedent layer of each low-level FNN. ADMM has proven to be an efficient solution for distributed computation problems \cite{boyd2011distributed,wang2019global}. We explored many algorithms designed for HFNNs but, to our best knowledge, an HFNN method with privacy preservation for heterogeneous big data has not yet been theorized or developed. Therefore, the contributions of this paper are three-fold:

\begin{itemize}
  \item We propose a new privacy-preserving HFNN model (PP-HFNN) to address some of the inherent issues with combining big heterogeneous data with machine learning tasks:  massive scales, heterogeneity, uncertainty, and privacy concerns.
  \item Privacy is preserved by using a distributed clustering method to identify the {\color{black}parameters in} the antecedent layer of each low-level FNN. This approach is also good at handling massive amounts of data.
  \item We have also developed a two-stage optimization algorithm to train the HFNN. Parameter learning for the low-level FNNs is based on the well-known ADMM, while coordination at the higher levels of the hierarchy relies on the AO method. The algorithm is scalable, fast, and does not suffer gradient vanishing problems, as is the case with methods based on back-propagation.
\end{itemize}

The remainder of the paper is organized as follows. In Section II, we {\color{black}discuss relevant} works on heterogeneous data, hierarchical fuzzy neural networks and distributed machine learning algorithms. Section III is devoted to the PP-HFNN modeling. In Section IV, we present the two-stage optimization algorithm, which consists of a distributed clustering method for the low-level FNNs and an AO method for high-level coordination. The results of a comprehensive simulation for both regression and classification tasks are presented and discussed in Section V to confirm the effectiveness of the proposed privacy-preserving HFNN model, and Section VI concludes the paper.

\nomenclature[S]{$\mathcal{N}$}{The set of samples.}
\nomenclature[S]{$\mathcal{F}$}{The set of features.}
\nomenclature[S]{$\mathcal{F}_b$}{The subset of features of the $b$-th low level FNN.}
\nomenclature[S]{$\mathcal{N}_{b,k}$}{The subset of training data assigned to the $k$-th agent of the $b$-th low-level FNN.}
\nomenclature[S]{$\mathcal{C}^{b}_k$}{The $k$-th cluster in the $b$-th low level FNN.}
\nomenclature[N]{$i$}{Notation for the sample.}
\nomenclature[N]{$j$}{Notation for the feature.}
\nomenclature[N]{$b$}{Notation for the low-level FNN.}
\nomenclature[N]{$l$}{Notation for the agent.}
\nomenclature[N]{$k$}{Notation for the cluster and the fuzzy rule.}
\nomenclature[N]{$x^{b,k}_{i,j}$}{The $j$-th component of the $i$-th sample assigned to the $k$-th agent of the $b$-th low-level FNN.}
\nomenclature[N]{$A_{r,j}^{b}$}{The $j$-th fuzzy set of the $r$-th rule on the $b$-th low-level FNN.}
\nomenclature[N]{$\mathbf{m, \sigma}$}{Notation for the parameters in the low level of the hierarchy.}
\nomenclature[N]{$\mathbf{w, v}$}{Notation for the parameters in the high level of the hierarchy.}
\nomenclature[A]{FNN}{Fuzzy neural network.}
\nomenclature[A]{HFNN}{Hierarchical fuzzy neural network.}
\nomenclature[A]{PP-HFNN}{Privacy-preserving hierarchical fuzzy neural network.}
\nomenclature[A]{ADMM}{Alternating direction method of multipliers.}
\nomenclature[A]{AO}{Alternating optmization.}
\begin{table*}
\begin{framed}
    \begin{multicols}{2}
     \printnomenclature
     \end{multicols}
\end{framed}
\end{table*}


{\color{black}
\section{Related Work}
This work is the first to combine privacy preservation with an HFNN designed for heterogeneous big data and a distributed computing environment. Therefore, in this literature review, we briefly cover relevant work in the three critical elements of PP-HFNN: heterogeneous data, HFNNs and distributed machine learning algorithms.

\subsection{Heterogeneous data}
As a way to resolve conflicts in heterogeneous data merged from a variety of sources,  Li et al. proposed an optimization framework to minimize the overall deviation between ground truths and the observations provided \cite{li2014resolving}. Zhang et al.’s solution combined principal component analysis (PCA) to reduce the number of dimensions with correlation analysis to investigate the interactive relations between features\cite{zhang2018research}. However, important information can be lost with this strategy, as PCA and correlation analysis do not consider nonlinear representative and interactive relations between features. A method based on support vector machine (SVM) was developed by Lewis et al. to infer functional gene annotations from heterogeneous data comprising protein sequences and structures \cite{lewis2006support}. Chen et al. developed a stacked denoising autoencoder to learn feature representations from hierarchical human mobility data. With this approach, they were able to predict the risk of traffic accidents \cite{chen2016learning}. Zuo et al. \cite{zuo2018fuzzy} proposed a fuzzy heterogeneous method to address domain adaptation with heterogeneous data spaces. However, none of the above methods \cite{li2014resolving,zhang2018research,lewis2006support,chen2016learning,zuo2018fuzzy} are a suitable solution to the uncertainty, scale issues and {\color{black}the} privacy concerns associated with big data environments.

\subsection{Hierarchical fuzzy neural networks}
Over the past decade, {\color{black}a range of} hierarchical structures of FNN have been studied for their benefits to different applications\cite{juang2007hierarchical,mohammadzadeh2013two,krlevza2016graph,deng2016hierarchical}. A hierarchical singleton-type recurrent FNN was proposed in \cite{juang2007hierarchical} for noisy speech recognition, consisting of two {\color{black} singleton-type} recurrent FNNs: one responsible for noise filtering and the other for recognition. Mohammadzadeh et al. \cite{mohammadzadeh2013two} proposed a hierarchical interval Type-2 FNN designed to synchronize uncertain chaotic systems. Searching for a
supervised framework more resilient to noisy inputs than the classical non-fuzzy neural network,
Krle{\v{z}}a et al. \cite{krlevza2016graph} proposed a fuzzy graph neural network based on a combination of fuzzy logic and {\color{black}a} recursive neural network for graph-matching tasks. Deng et al. \cite{deng2016hierarchical} developed a hierarchical network structure by fusing a conventional FNN and a deep neural network. {\color{black}Benefiting} from fuzzy logic, this method is able to increase classification accuracy with data that contains a high level of uncertainty. {\color{black}However,} these hierarchical FNNs \cite{juang2007hierarchical,mohammadzadeh2013two,krlevza2016graph,deng2016hierarchical}  cannot handle heterogeneous big data effectively or efficiently, and some still suffer from the curse of dimensionality concerning fuzzy rules. Unlike these approaches, our PP-HFNN framework is designed to extract features from heterogeneous data and reduce the number of fuzzy rules without compromising performance.

\subsection{Distributed machine learning algorithms}
Distributed machine learning algorithms have been widely investigated\cite{forero2010consensus,qin2016distributed,bi2015distributed,ye2019decentralized,scardapane2016decentralized}. For instance, Forero et al. \cite{forero2010consensus} developed a consensus-based distributed SVM for a multi-agent network, where the training data is distributed across different agents but {\color{black}data exchange} among agents is prohibited. Qin et al. \cite{qin2016distributed} proposed a distributed K-means algorithm and a distributed fuzzy c-means model for wireless sensor networks based on distributed consensus strategies. A distributed extreme learning machine with kernels based on MapReduce was proposed by Zhao et al. \cite{bi2015distributed}, designed for very fast learning speeds. Ye et al.\cite{ye2019decentralized} developed a decentralized multi-task extreme learning machine based on a hybrid Jacobian and Gauss-Seidel proximal multi-block ADMM method. A decentralized training algorithm for Echo State Networks was developed by Scardapane et al. \cite{scardapane2016decentralized} directed toward distributed big data applications. {\color{black}But,} although these distributed machine learning algorithms \cite{forero2010consensus,qin2016distributed,bi2015distributed,ye2019decentralized,scardapane2016decentralized} can reduce computational requirements and alleviate potential privacy concerns, they may not be effective strategies for dealing with data uncertainty and heterogeneity.

In the last few years, several distributed algorithms for FNN have been proposed  \cite{fierimonte2016distributed,fierimonte2017distributed,ye2020consensus}. Fierimonte et al. presented a decentralized algorithm for a random-weight FNN, where the parameters in the antecedent layer are randomly selected instead of being estimated \cite{fierimonte2016distributed}. Subsequent work saw an online implementation of the same FNN structure in \cite{fierimonte2017distributed}. However, a random method of identifying parameters may result in very large deviations of accuracy during the learning process.  In addition, this approach suffers from the curse of dimensionality {\color{black}because} the number of fuzzy rules increases exponentially as the input space increases. Moreover, the distributed algorithms are only applied {\color{black}to the} consequent layers of the FNN. To {\color{black}retrieve} the parameters of the antecedent layer, all the data must first be processed by a central agent. A further issue is that this method does not provide any level of privacy protection, and the FNNs proposed can neither handle the scale of big data nor any heterogeneity issues.
Recently, a consensus learning method for distributed FNN was proposed in \cite{ye2020consensus}, where a distributed clustering method and a distributed parameter learning method were respectively developed to locate the parameters in the {\color{black}antecedent and consequent layers of an FNN}. Yet, again, none of the above distributed algorithms for FNN \cite{fierimonte2016distributed,fierimonte2017distributed,ye2020consensus} can address the issue of data heterogeneity in big data environments.
}

\section{Model Formulation of the PP-HFNN for Heterogeneous Big Data}
\begin{figure*}[!htbp]
  \centering
  \includegraphics[width=1.8\columnwidth]{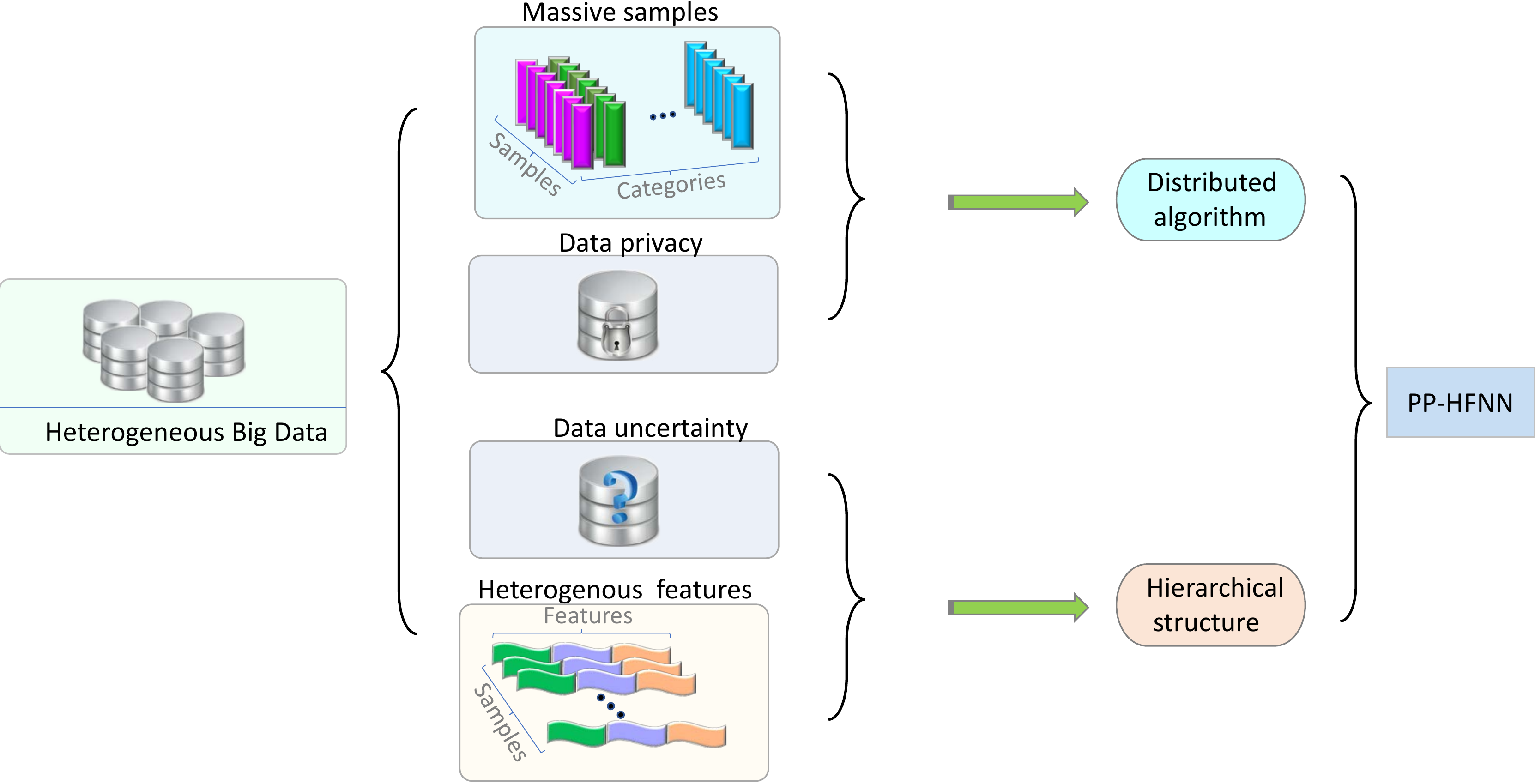}
  \vspace{-10pt}
  \caption{\normalsize {\color{black}The challenges with heterogeneous big data and the motivation behind PP-HFNN}}
  \label{fig:contribution}
\end{figure*}

Heterogeneous big data is inherently characterized by its enormous scale, heterogeneity and uncertainty and is often interspersed with sensitive personal data. HFNN, which takes advantage of the hierarchical structure and fuzzy if-then rules, is effective for tackling the challenges of data heterogeneity and uncertainty. {\color{black}But the issues of enormous scale and data privacy fall into the domain of distributed privacy-preserving algorithms. As shown in Fig. \ref{fig:contribution}, these challenges motivated our PP-HFNN model as a promising solution.}

\begin{figure*}[!t]
  \centering
  \includegraphics[width=1.3\columnwidth]{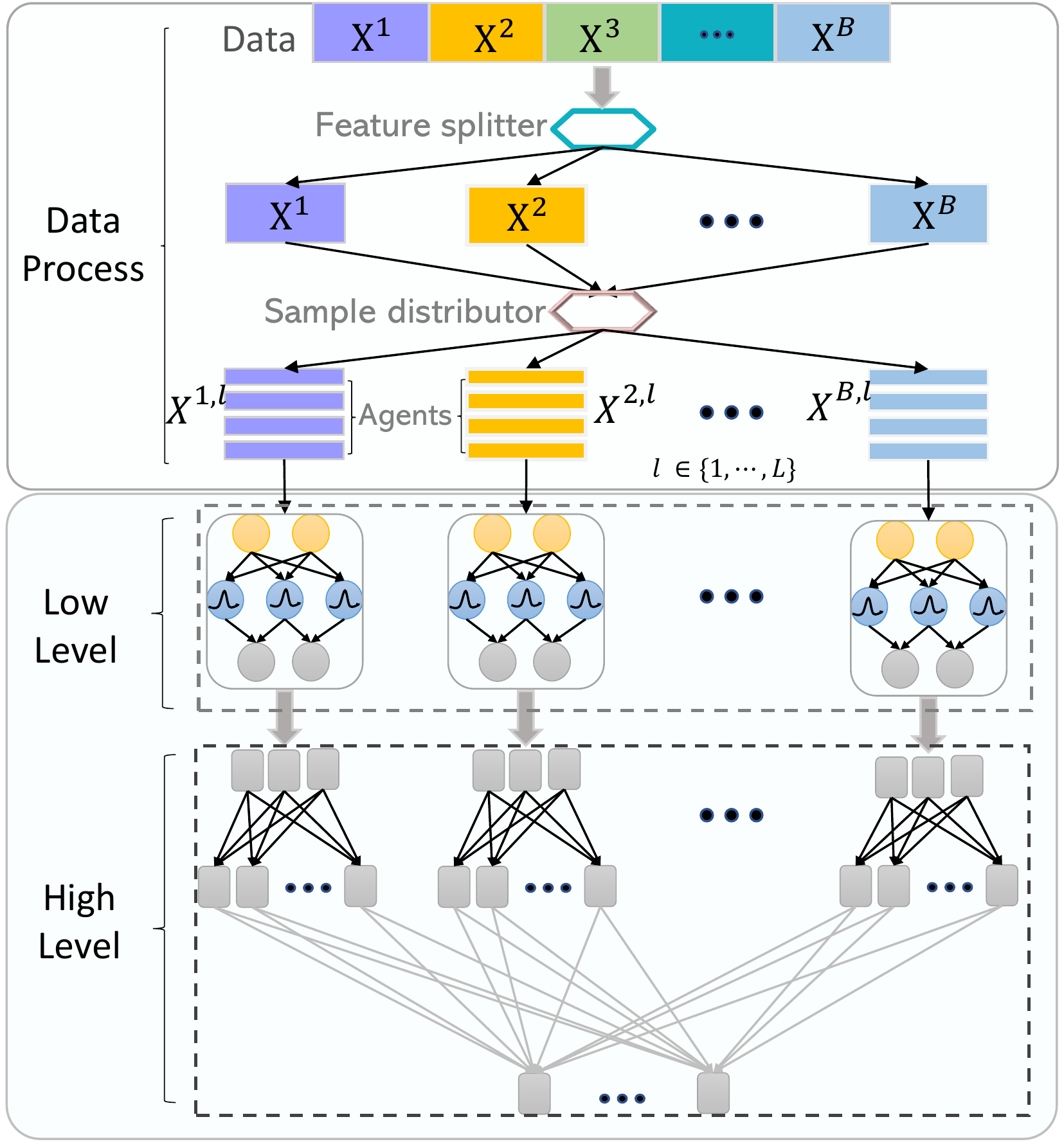}
  \vspace{-10pt}
  \caption{\normalsize Data processing and PP-HFNN structure.}
  \label{fig:stucture}
\end{figure*}

PP-HFNN consists of {\color{black}a two-level} hierarchical structure. The {\color{black}lower level contains multiple FNNs, and the outputs of these FNNs are  in the higher level.} During the data processing stage, heterogeneous big data is segmented {\color{black}at both} the feature level and the sample level. In the feature-level segmentation, the heterogeneous features are assigned {\color{black}to} multiple subsets by a feature splitter to ensure that each subset {\color{black}only} contains homogeneous features. {\color{black}Further,} each low-level FNN is responsible for only one subset. {\color{black}The samples in the sample-level segmentation for each low-level FNN are further randomly assigned to multiple agents.} A distributed computation framework is then applied to these agents in each low-level FNN to preserve data privacy and {\color{black}to deal with the enormous amounts of data samples.} The above data processing and PP-HFNN structure is shown in Fig.\ref{fig:stucture}.

Suppose $\clD:=\{(X_i,Y_i)|\ X_i\in \mathbb{R}^{|\clF|}, Y_i\in \mathbb{R}, i\in \clN \}$ denote the set of heterogeneous data, where $\clF$ and $\clN$ represents the set of features and samples, respectively, and $|\cdot|$ is the cardinality operator. As shown in Fig.\ref{fig:stucture}, the data is first processed {\color{black}by a feature splitter where the full set of heterogeneous features is transformed into multiple independent subsets of homogeneous features and assigned to a low-level FNN.} Let $\clF_b, b\in \{1,2,...,B\}$ denote the subset of homogeneous features associated with {\color{black}the} $b$-th low-level FNN. {\color{black}Then,} in each low-level FNN, the training data is further segmented by a sample distributor. $\clN_{b,l}$ {\color{black}denotes} the subset of training data assigned to the $l$-th agent ($l\in \{1,2,...,L\}$) of the $b$-th low-level FNN. Accordingly, the sample vector is collected in $X^{b,l}_{i}$, $\forall i\in \clN_{b,l}$ with $x^{b,l}_{i,j}$ {\color{black}denoting} its $j$-th component.

Next, let us briefly recall the structure of {\color{black}the} low-level FNNs, which {\color{black}follows the first-order of the Takagi-Sugeno method for fuzzy inference systems} and has a layer-by-layer network structure  \cite{lin1991neural}. {\color{black}Now,} suppose the output of the $l$-th agent of the $b$-th low-level FNN is $z_b^l$, then the $k_b^l$-th fuzzy rule can be {\color{black}expressed} as
\begin{center}
  Rule $k_b^l$: IF $x_{i,1}^{b,l}$ is $A_{r,1}^{b,l}$, $\cdots$ and $x_{i,|\mathcal{F}_b|}^{b,l}$ is $A_{r,|\mathcal{F}_b|}^{b,l}$\\
    Then $z_{r_b^l}$ = $w_{k,0}^{b,l} + \sum_{j=1}^{|\mathcal{F}_b|}w_{k,j}^{b,l}x_{j,1}^{b,l}$,
\end{center}
where $A_{k,j}^{b,l}$ is a Gaussian fuzzy set of the $j$-th input of rule $k_b^l$, and $w_{k,j}^{b,l}$ is the corresponding weight of the consequence. The membership function of $A_{k,j}^{b,l}$ can be written as
\begin{equation}\label{Gaussian_A}
\varphi_{k,j}(x_{i,j}^{b,l}) = \mbox{exp}\left[-\left(\frac{x_{i,j}^{b,l}-m_{k,j}^{b,l}}{\sigma_{k,j}^{b,l}}\right)^2\right]
\end{equation}
where $m_{k,j}^{b,l}$ and $\sigma_{k,j}^{b,l}$ are respectively the center and width of the corresponding fuzzy set. Generally, each low-level FNN {\color{black}is composed of} four feed-forward layers, as shown on the right-hand side of Fig.\ref{fig:stucture}.

Layer 1 is the antecedent layer: its inputs are $x_{i,j}^{b,l}$, and its outputs are {\color{black}derived through the fuzzification process outlined in} (\ref{Gaussian_A}).

Layer 2 is the rule layer: each node in this layer represents a fuzzy rule, which {\color{black}uses} the AND operation to match the outputs of {\color{black}the} antecedent layer as follows:
\begin{equation}\label{firing}
\phi_k(X_{i}^{b,l}) = \prod_{j=1}^{|\mathcal{F}_b|}\varphi_{k,j}(x_{i,j}^{b,l}),
\end{equation}
{\color{black}where $\phi_k(X_{i}^{b,l})$ is the firing strength of fuzzy rule $k_b^l$, and then normalized by}
\begin{equation}\label{firing_norm}
\bar{\phi}_k(X_{i}^{b,l}) = \frac{\phi_k(X_{i}^{b,l})}{\sum_{k=1}^{K^b_l}\phi_r(X_{i}^{b,l})}.
\end{equation}
where $K^b_l$ is the total number of fuzzy rules in the $l$-th agent of the $b$-th low-level FNN.

Layer 3 is the consequent layer: each node here performs a defuzzification process for each fuzzy rule $r_b^l$ using a weighted average operation as follows:
\begin{equation}\label{defuzzy}
\psi_k(X_{i}^{b,l}) = \bar{\phi}_k^{b,l}(X_{i}^{b,l})(w_{k,0}^{b,l} + \sum_{j=1}^{|\mathcal{F}_b|}w_{k,j}^{b,l}x_{j,1}^{b,l}),
\end{equation}

Layer 4 is the output layer: the overall output of the $l$-th agent of the $b$-th low-level FNN is {\color{black}derived by summing the outputs of the fuzzy rules in Layer 3 as follows:}
\begin{equation}\label{layer4}
z_i^{b,l} = \sum_{k=1}^{K^b_l}\psi_k^{b,l}(X_{i}^{b,l}).
\end{equation}

For each $l$-th agent of the $b$-th low-level FNN, a matrix {\color{black}is defined as} $H^{b,l}:= [H_1^{b,l},\cdots, H_K^{b,l}]$, where $H_k^{b,l}\in \mathbb{R}^{|\clN|\times (|\clF_b|+1)}$, and
\[
H_k^{b,l}=\begin{bmatrix}
\bar{\phi}_k(X_{1}^{b,l}) & \bar{\phi}_k(X_{1}^{b,l})x_{1,1}^{b,l} & \cdots & \bar{\phi}_k(X_{1}^{b,l})x_{1,|\clF_b|}^{b,l} \\
\bar{\phi}_k(X_{2}^{b,l}) & \bar{\phi}_k(X_{2}^{b,l})x_{2,1}^{b,l} & \cdots & \bar{\phi}_k(X_{2}^{b,l})x_{2,|\clF_b|}^{b,l} \\
\vdots & \vdots & \ddots & \vdots \\
\bar{\phi}_k(X_{|\clN|}^{b,l}) & \bar{\phi}_k(X_{1}^{b,l})x_{|\clN|,1}^{b,l} & \cdots & \bar{\phi}_k(X_{1}^{b,l})x_{|\clN|,|\clF_b|}^{b,l}
\end{bmatrix}
\]
The output vector is \[Z^{b,l}:=[z_1^{b,l},\cdots,z_{|\clN|}^{b,l}]^T \in \mathbb{R}^{|\clN|}.\]
And the weight vector is
\[
\bw^{b,l}: = [w_{1,0}^{b,l},\cdots,w_{1,|\clF_b|}^{b,l},\cdots,w_{K_l^b,0}^{b,l},\cdots,w_{K_l^b,|\clF_b|}^{b,l}]^T.
\]
Eq. (\ref{layer4}) is equivalent to the following matrix form:
\begin{equation}\label{layer4_matrix}
Z^{b,l} = H^{b,l}\bw^{b,l}.
\end{equation}
{\color{black}It is} worth noting that $H^{b,l}$ is dependent on the centers and widths of each rule's fuzzy sets, i.e., $m_{k}^{b,l}$ and $\sigma_{k}^{b,l}$ {\color{black}as well as} the data input $X_i^{b,l}$.

{\color{black}The outputs of the low-level FNNs are coordinated by a fully-connected layer in the high level of the hierarchy.} For each agent $l$,
\begin{equation}\label{coordination}
Y^l = \sum_{b=1}^{B} Z^{b,l} v^{b,l},
\end{equation}
where $Y^l:=[Y_1^l,\cdots,Y_{\clN}^l]$, $v^{b,l}$ is the coordination weight of the hierarchy, and $B$ is the number of subsets in terms of homogeneous features. Let $Z^l$ {\color{black}represent the collection of outputs from} each low-level FNN of the $l$-th agent, i.e.,
\[
Z^l:=[Z^{1,l},\cdots,Z^{B,l}]\in \mathbb{R}^{|\clN|\times B}.
\]
Eq. (\ref{coordination_matrix}) is equivalent to the following matrix form:
\begin{equation}\label{coordination_matrix}
Y^l = Z^l\bv^l,
\end{equation}
where \[\bv^l:=[v^{1,l},\cdots,v^{B,l}]^T \in \mathbb{R}^B.\]
Define
\[\bm^l:=[\bm^{1,l},\cdots,\bm^{B,l}],\]
\[\mathbf{\sigma}^l:=[\mathbf{\sigma}^{1,l},\cdots,\mathbf{\sigma}^{B,l}],\]
\[H^l:=[H^{1,l},\cdots,H^{B,l}] \in \mathbb{R}^{|\clN|\times n_H},\]
and
\[
\bw^l:=\mbox{diag}(w^{1,l},\cdots,w^{B,l})\in \mathbb{R}^{n_H\times B},
\]
where $n_H = \sum_{b=1}^{B}K_l^b(|\clF_b|+1)$, and $\mbox{diag}(\cdot)$ represents {\color{black}a diagonal operation on the matrix}. The training procedure {\color{black}for} PP-HFNN is to solve the following optimization problem:
\begin{equation}\label{optimization1}
  \ds\min_{\bm^l,\mathbf{\sigma}^l,\bw^l,\bv^l} \frac{1}{2}\sum_{l=1}^{L}(||Y^l- H^l \bw^l \bv^l||^2 + \frac{\lambda}{2}||\bw^l||^2 + \frac{\mu}{2}||\bv^l||^2),
\end{equation}
where $L$ is the number of agents, $\lambda$ and $\mu>0$ are factors to {\color{black}trade-off}  the training error and regularization.
{\color{black}Selecting an appropriate value for $\lambda$ and $\mu>0$ can make the solution much more stable and generalizable}\cite{huang2011extreme}.
The difficulty of the optimization problem (\ref{optimization1}) {\color{black}stem from three issues. One is the nonconvex nature} of the Gaussian membership function in $H^l$. To address this {\color{black}first difficulty}, the key issue is to identify the parameters of the centers $\bm^l$ and {\color{black}the} widths $\mathbf{\sigma}^l$ of each rule's fuzzy sets in the antecedent {\color{black}layer} of each FNN. The second difficulty is the {\color{black} distributed computing scheme for the low-level FNNs, which is needed to manage the scale of the data and preserve privacy. The last difficulty is the} coordination in the high levels of the hierarchy, where the matrix product between $\bw^l$ and $\bv^l$ is nonconvex. Fortunately, it is convex after fixing either $\bw^l$ or $\bv^l$, i.e., it {\color{black}becomes} bi-convex.

\section{Two-stage optimization algorithm to train the PP-HFNN}

An intuitive way to solve the nonconvex optimization (\ref{optimization1}) is by using a back-propagation method\cite{juang1998online}. However, {\color{black}back-propagation methods} often suffer from slow training speeds and gradient vanishing problems. In addition, distributed computation is not {\color{black}easily implemented with} back-propagation. {\color{black}Hence, it is not an optimal choice for heterogeneous big data environments. A more suitable option is an algorithm that is fast to converge with a massive number of samples and is easy to roll out across a distributed computation framework.}

As {\color{black}discussed} above, the first difficulty with the optimization problem (\ref{optimization1}) is to identify the parameters of the antecedent layer of each FNN. A simple idea is to group the input data into multiple clusters and use one rule for each cluster \cite{chiu1994fuzzy}. Here, we use the popular K-means algorithm\cite{macqueen1967some} to identify the parameters of the antecedent layer. The K-means algorithm is one of the most efficient clustering algorithms.
To tackle the second difficulty of the optimization problem (\ref{optimization1}), {\color{black}we developed} a distributed K-means method, inspired by \cite{forero2011distributed}. As for the coordination in high level coordination of the hierarchy, the AO method is used since it is well-suited to the bi-convex optimization problems and converges very quickly in practice.


\subsection{Distributed K-means method for the low-level of HFNN}
To introduce the distributed K-means algorithm, let us first recall the centralized K-mean algorithm. {\color{black}The goal of this algorithm is to assemble the input data into multiple clusters $\mathcal{C}^{b}_k$,} each of which has its own center. The centralized K-means algorithm for the $b$-th low-level FNN of the hierarchy can be defined {\color{black}as}
\begin{equation}\label{cluster1}
 \bm^{b}_k = \mbox{arg}\ds\min_{\bm^{b}_k} \frac{1}{2} \sum_{k=1}^{K_b}\sum_{X_i^b\in \clC_k^b}||X^{b}_i-\bm^{b}_k ||^2
\end{equation}
where $\bm^{b}_k$ is the $k$-th center in the antecedent layer of the $b$-th low-level FNN, and $K_b$ {\color{black}represents} the number of corresponding clusters, which is equal to the number of rules. Starting from an initial set of $K$ centers, i.e. $\{\bm_1^b(0),\cdots,\bm_K^b(0)\}$, the {\color{black}K-means} algorithm alternates between an assignment step and an update step as follows.
\begin{itemize}
  \item Assign each observation $X_i^b$ to the cluster $\mathcal{C}_k^b(t)$, whose center is closest to $X_i^b$.
  \item Update the center of each cluster by \[\bm_k^b(t) = \frac{1}{|\mathcal{C}_k^b(t)|}\ds\sum_{X_i^b\in \mathcal{C}_k^b(t)}X_i^b.\]
\end{itemize}
Once the center of each fuzzy set is obtained by the above procedure, the corresponding  standard variance $\sigma_{k,j}^b$ can be calculated as follows:
\begin{equation}\label{standard_v}
  \sigma_{k,j}^b = \sqrt{\frac{1}{|\mathcal{C}_k^b(t)|}\ds\sum_{i=1}^{|\mathcal{C}_k^b(t)|} (x_{i,j}^b-m_{k,j}^b)^2}.
\end{equation}

\begin{figure*}[!htbp]
  \centering
  \includegraphics[width=1.8\columnwidth]{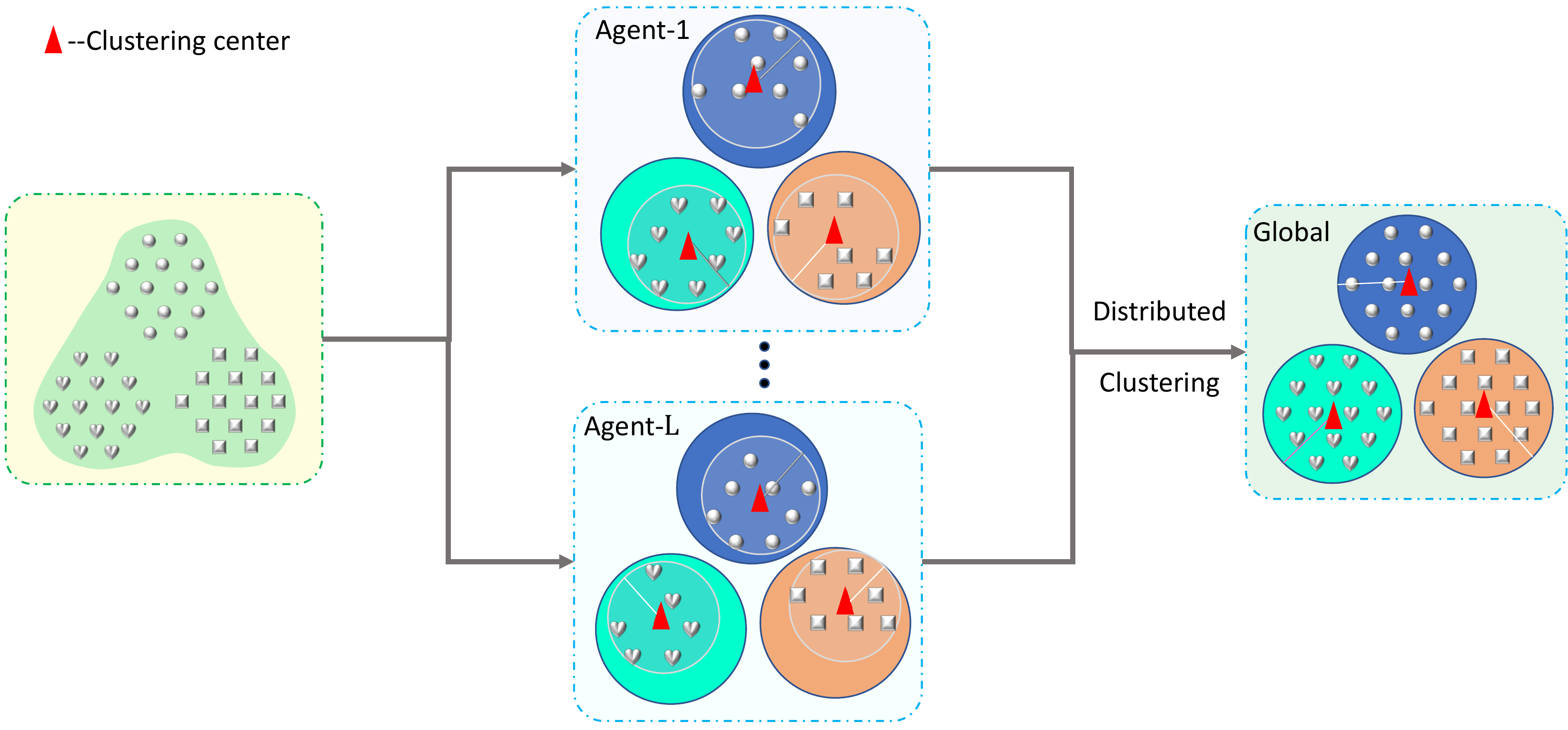}
  \vspace{-10pt}
  \caption{\normalsize Distributed Clustering.}
  \label{fig:admm_kmm}
\end{figure*}
It is clear that such a centralized K-means algorithm {\color{black}does not preserve privacy at all, but the distributed K-means algorithm does}. As shown in Fig. \ref{fig:admm_kmm}, the data is processed locally by multiple agents. {\color{black}While some limited information is exchanged between agents, none of it is data.} The distributed K-means is built on ADMM, which is an efficient solution for distributed computation. A recent study proves the convergence of ADMM for a variety of nonconvex and possibly nonsmooth functions given some sufficient conditions\cite{wang2019global}. Following our previous work\cite{ye2020consensus}, the distributed K-means algorithm solves the following optimization problem:
\begin{subequations}\label{global_m}
\begin{eqnarray}
   \ds\min_{\bm^l_K} \frac{1}{2} \sum_{l=1}^{L}\sum_{b=1}^{B}\sum_{X_i^b\in \clC^b_k}||X^{b,l}_{i}-\bm^{b,l}_k||^2\label{str_a}\\
   \mbox{s.t.}\quad  \bm^{b,l}_k = \mathbf{r}_k^b, \quad l=1,2,\cdots,L \label{str_b}
\end{eqnarray}
\end{subequations}
where $\bm^{b,l}_k$ represents the $k$-th local center of the $l$-th agent on the $b$-th low-level FNN, and $\mathbf{r}_k^b$ is the corresponding global center. The global standard variance of the antecedent layer of the low-level FNNs is {\color{black}expressed} as
\begin{equation}\label{global_std}
  \bar{\sigma}^{b}_k = \sqrt{\frac{1}{|\mathcal{N}|}\ds\sum_{l=1}^{L}|\mathcal{C}^{b,l}_k|(\sigma^{b,l}_{k})^2}
\end{equation}
where $\bar{\sigma}^{b}_k$ denotes the $k$-th global standard variance of the $k$-th low-level layer, and $\sigma^{b,l}_{k}$ is the local standard variance of the $l$-th agent.

The following augmented Lagrangian is contructed for
(\ref{global_m}):
\begin{eqnarray}\label{alm_distribute_m}
  \clL(\bm^{b,l}_k,\br^b_k,\boldsymbol{\beta}_{k}^{b,l}) = && \frac{1}{2} \sum_{l=1}^{L}\sum_{k=1}^{K}\sum_{X_i^b\in \mathcal{C}^{b,l}_k}||X^{b,l}_{i}-\bm^{b,l}_{k}||^2 \nonumber\\ && + \ds\sum_{l=1}^{L}\sum_{k=1}^{K}{\boldsymbol{\beta}_{k}^{b,l}}^T(\bm^{b,l}_k - \br^{b,l}_k) \nonumber\\ && +
 \frac{1}{2}\rho \ds\sum_{l=1}^{L}\sum_{k=1}^{K}||\bm^{b,l}_k - \br^{b,l}_k||^2
\end{eqnarray}
where $\boldsymbol{\beta}_{k}^{l}$ is the Lagrange multiplier, and $\rho$ is a positive penalty parameter. The variables are then updated iteratively with the following procedure based on the ADMM:
\begin{eqnarray}
\bm^{b,l}_{k}(t+1) = \mbox{arg}\ds\min_{\bm}\clL(\bm^{b,l}_{k},\br^{b}_k(t),\boldsymbol{\beta}_{k}^{b,l}(t)),\label{ADMM_s1}\\
\br^{b}_k(t+1) = \mbox{arg}\ds\min_{\br^{b,l}_k}\clL(\bm^{b,l}_{k}(t+1),\br^{b}_k,\boldsymbol{\beta}_{k}^{b,l}(t)),\label{ADMM_s2}\\
\boldsymbol{\beta}_{k}^{b,l}(t+1) = \boldsymbol{\beta}_{k}^{b,l}(t) + \rho\ds\sum_{l=1}^{L}\sum_{k=1}^{K}(\bm^{b,l}_{k}(t+1) - \nonumber\\ \br^{b}_k(t+1)),\label{ADMM_s3}
\end{eqnarray}
where $t$ is the number of {\color{black}iterations}. It is worth noting that the distributed K-means algorithm is applied {\color{black}to} each low-level FNN and the centers $\bm^{b,l}_{k}$ can be updated in parallel by different agents. {\color{black}Additionally,} there is a closed-form solution for (\ref{ADMM_s2}) as follows:
\begin{eqnarray}\label{close_r}
\br^{b,l}_k(t+1) = \frac{1}{L\rho} \sum_{l=1}^{L}(\boldsymbol{\beta}_{k}^{b,l}(t)+\rho\bm^{b,l}_{k}(t+1))
\end{eqnarray}
{\color{black}Convergence depends on} the following two criteria:
\begin{eqnarray}\label{converg_alg1}
||\bm^{b,l}_{k}(t) - \br^{b}_k(t)||^2 \leq \epsilon_1,\\
||\boldsymbol{\beta}_{k}^{b,l}(t+1) - \boldsymbol{\beta}_{k}^{b,l}(t)||^2 \leq \epsilon_2.
\end{eqnarray}

\subsection{Alternating optimization for the high-level of HFNN}
In the second stage, the AO method is used to obtain the parameters $\bw$ and $\bv$ in high levels of the hierarchy. Since the raw data is only processed in the first stage, it is not necessary to use distributed computation for the high-level coordination from {\color{black}a} privacy-preserving perspective. Therefore,
the optimization problem (\ref{optimization1}) can be simplified to:
\begin{equation}\label{optimization2}
  \ds\min_{\bw,\bv} \frac{1}{2}||Y - H \bw \bv||^2 + \frac{\lambda}{2}||\bw||^2 + \frac{\mu}{2}||\bv||^2,
\end{equation}
which does not incorporate the agents. Obviously, (\ref{optimization2}) is a bi-convex optimization problem. After fixing either $\bw$ or $\bv$, {\color{black} it becomes convex. As mentioned}, the AO method is a good choice for this bi-convex optimization problem. {\color{black}The process is as follows:}

Update $\bw$ with fixed $\bv(t)$:  Let $\hat{\bw}$ {\color{black}be the collection of} $\{\bw^1,\cdots,\bw^B\}$ in vector form. By introducing a matrix $H_w(t) := [v^1(t)H^1,\cdots,v^B(t)H^B]$, the weight $\hat{\bw}$ can be {\color{black}calculated} by solving the following optimization problem:
\begin{equation}\label{weight_w}
  \hat{\bw} = \mbox{arg}\ds\min_{\hat{\bw}} \frac{1}{2}||Y-H_w(t)\hat{\bw}||^2 + \frac{\lambda}{2}||\hat{\bw}||^2,
\end{equation}
which is a standard {\color{black}least-squares} optimization problem. Its closed-form solution can be by setting its partial derivative to 0, i.e.,
\begin{equation}\label{solution_w}
  \hat{\bw}(t) = (H_w^TH_w + \lambda I)^{-1}H_w^TY.
\end{equation}

Update $\bv$ with fix $\bw(t)$: The hidden output vector $Z(t)$ of the low-level FNNs is {\color{black}generated} after $\bw(t)$ is updated. Then the weight $\bv(t)$ can be updated by
\begin{equation}\label{weight_v}
  \bv = \mbox{arg}\ds\min_{\bv} \frac{1}{2}||Y-Z(t)\bv||^2 + \frac{\mu}{2}||\bv||^2,
\end{equation}
Similarly, (\ref{weight_v}) can be solved through:
\begin{equation}\label{solution_v}
  \bv(t) = (Z(t)^TZ(t) + \mu I)^{-1}Z(t)^TY,
\end{equation}

A summary of the two-stage PP-HFNN optimization algorithm is provided in Algorithm \ref{alg1}.
\begin{algorithm}
\caption{Two-stage optimization algorithm for PP-HFNN} \label{alg1}
\begin{algorithmic}
\STATE{\textbf{Stage-1: ADMM-based distributed clustering (\ref{global_m})}}
\STATE{\textbf{Initialization:} Set $t=0$ and randomly initialize the global cluster center $\br^{b}_k(t)$ and Lagrange multipliers $\boldsymbol{\beta}_{k}^{b,l}(t)$ for each agent in {\color{black}the} low-level layers, and randomly initiate the local cluster centers $\bm^{b,l}_{k}$ for all agents {\color{black}in} all low-level layers.}
\FOR{$t=0,1,2,\cdots,$}
\STATE {\textbf{Update the local center $\bm^{b,l}_{k}(t+1)$:}
\\ Assignment step: Each agent of each low-level layer assigns its data $X_i^{b,l}$ to the cluster $\clC_k^{b,l}(t-1)$, derived in the previous iteration. \\
Update step: Each agent $b_l$ updates the center of each cluster $\clC_k^{b,l}(t)$ by
\begin{equation}\label{update_local}
 \bm^{b,l}_{k}(t+1)=\frac{1}{|\mathcal{C}_k^{b,l}(t)|}\ds\sum_{X_i^{b,l}\in \mathcal{C}_k^{b,l}(t)}X_i^{b,l}
\end{equation} }
\STATE {\textbf{Update the global variables $\br^{b}_k(t+1)$} by  (\ref{close_r}) and broadcast it to each agent $l$. }
\STATE {\textbf{Update the dual variables $\boldsymbol{\beta}_{k}^{b,l}(t+1)$} by (\ref{ADMM_s3}) and broadcast it to each agent $l$}
\ENDFOR

\STATE{\textbf{Stage-2: AO method for high-level layer (\ref{optimization1})}}
\STATE{\textbf{Initialization:}Set $t=0$, and randomly initialize the output weight $\bv$ and {\color{black}the} Lagrange multipliers $\boldsymbol{\lambda}$ and $\boldsymbol{\mu}$. Transform $\bw$ into $\hat{\bw}$}
\FOR{$t=0,1,2,\cdots,$}
\STATE {Fix $\bv$ and update \textbf{$\hat{\bw}$ }by (\ref{solution_w}).}
\STATE {Fix $\hat{\bw}$ and update \textbf{$\bv$ }by (\ref{solution_v}).}
\ENDFOR
\end{algorithmic}
\end{algorithm}

\section{Experimental Evaluation}
{\color{black} To evaluate the performance of the proposed PP-HFNN,  we first tested it on an artificial dataset and investigated how noise and outliers affected performance.} Then, we tested the PP-HFNN on three real datasets with heterogeneous features {\color{black}on both} regression and classification tasks. The datasets were: High Storage System Data (HRSS)  \cite{hranisavljevic2016novel}, EEG signals from the Distracted Driving experiment (EEG-DD)  \cite{lin2011eeg}, and Ethelen-CO (EC) \cite{fonollosa2015reservoir}. Since samples of each dataset were either recorded from different types of sensors or assembled from different sources, they are naturally heterogeneous. Details of these datasets are as follows.

The HRSS dataset consists of 6544 records of signal data in two categories captured by 6 sensors equipped to 6 belts of a high storage system. The system generates four datasets according to different belt moving conditions. The features collected by each sensor include transport distance, power, and belt voltage. Therefore, each sample contains 18  features (3 features for each of 6 sensors).

The  EEEG-DD dataset contains EEG signals from 11 subjects using 6 channels, which are transformed into 4 bands of frequency data for each channel. Each subject is required to complete two different tasks several times. The signals recorded each time a subject completes one task is called a trial, and each trial is divided into 17 segments. Further, each subject has a different number of trials, ranging from 96 to 183. Thus each segment has 24 features (6 channels divided into 4 bands).

The EC dataset contains time series data of Ethylene and CO readings in the air under dynamic gas mixtures. It contains 4,208,261 samples with 16 features recorded by 16 chemical sensors of 4 different types.

The details of these three datasets are summarized in Table I, where the second, third and fourth column provides the number of features, the low-level FNNs and the samples, respectively.
\begin{table}[!ht]
  \centering
  {\fontsize{9pt}{10pt}\selectfont
  \caption{Dataset and Parameter details}
  \begin{tabular}{|c|c|c|c|c|c|} \hline
  \multirow{2}{*}{Dataset} &\multirow{2}{*}{Feature}&\multirow{2}{*}{Brunch}&\multirow{2}{*}{Sample}& \multicolumn{2}{|c|}{Parameter Setting}   \\ \cline{5-6}
               &         &        &           & Agents & Rules               \\ \hline
   HRSS        & 18      & 6      & 6544      & 5      & 1-5                 \\ \hline
   EEG-DD      & 24      & 6      & -         & 5      & 1-15                \\ \hline
   EC          & 16      & 4      & 4,208,261 & 200    & 1-50                \\ \hline
  \end{tabular}}
  \vspace{-10pt}
  \label{datasets}
\end{table}

To make the simulation more {\color{black}realistic}, we conducted both classification and regression tasks. We used the HRSS and EEG-DD datasets for the classification task and the EC dataset with a very large number of  samples for the regression task. All classification and regression experiments were conducted on a laptop Intel i7 with a 4.0 GHz processor and 16 GB of memory. As listed in Table I, all the parameters mentioned in the above algorithms were selected by searching the spaces of $[10^{-5}, 10^{-4},\cdots, 10^{5}]$.
{\color{black}
\subsection{Artificial dataset}
The artificial dataset we built to investigate the impact of noise and outliers on PP-HFNN’s performance was generated as follows. The data input $X$ consisted of two branches of features to stimulate heterogeneous data, i.e., $X = {[{X^{(1)}}^T, {X^{(2)}}^T]}^T$, where
$X^{(1)} =[x^{(1)}_1, x^{(1)}_2, x^{(1)}_3]^T$ and $X^{(2)} =[x^{(2)}_1, x^{(2)}_2, x^{(2)}_3]^T$. The corresponding output $Y$ was given by:
\begin{equation}\label{AD}
  Y = 0.3\sum_{i=1}^{3}(\cos(x^{(1)}_i))^2 + 0.7\sum_{i=1}^{3}(\cos(x^{(2)}_i)),
\end{equation}
The feature inputs $X^{(1)}$ {\color{black}were} generated from a Gaussian distribution with a mean of $[1, 5, 9]$ and a standard deviation of $[0.1, 0.2, 0.3]$, while $X^{(2)}$ {\color{black}was generated from} a Gaussian distribution with a mean of $[3, 7, 2]$ and a standard deviation of $[0.2, 0.4, 0.1]$. We generated a total of 50,000 samples. Three different levels of noise, 5\%, 10\% and 15\%, and three different numbers of outliers, 1\%, 5\% and 10\%, were then added to the feature inputs $X$. Noise for the feature inputs was generated from a Gaussian distribution $\mathcal{N}(0.1R,(0.1R)^2)$, where $R = max(X) - min(X)$. Similarly, the outliers were generated from a Gaussian distribution $\mathcal{N}(R,(0.1R)^2)$. The data in each low-level branch of PP-HFNN {\color{black}were} separated and allocated to five different agents, and 10 fuzzy rules were used.
PP-HFNN's performance on the artificial dataset with different levels of noise and outliers is shown in Fig. \ref{fig:noise_analysis}. It is clear that it is quite robust to noise and outliers at various levels.
\begin{figure*}[!ht]
  \centering
  \subfigure{
  \includegraphics[width=0.9\columnwidth]{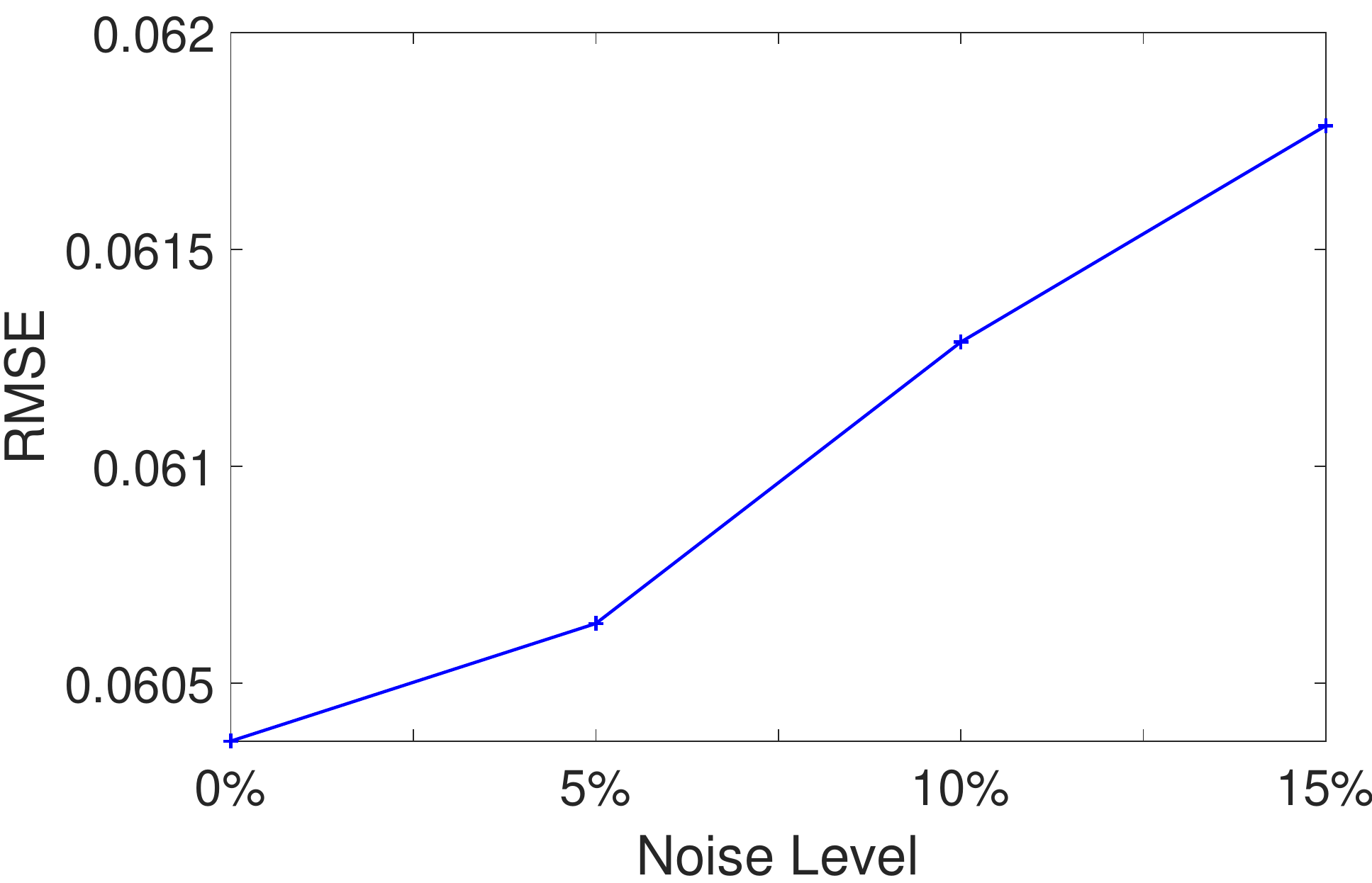}
  }
  \vspace{3pt}
  \hspace{3pt}
  \subfigure{
  \includegraphics[width=0.9\columnwidth]{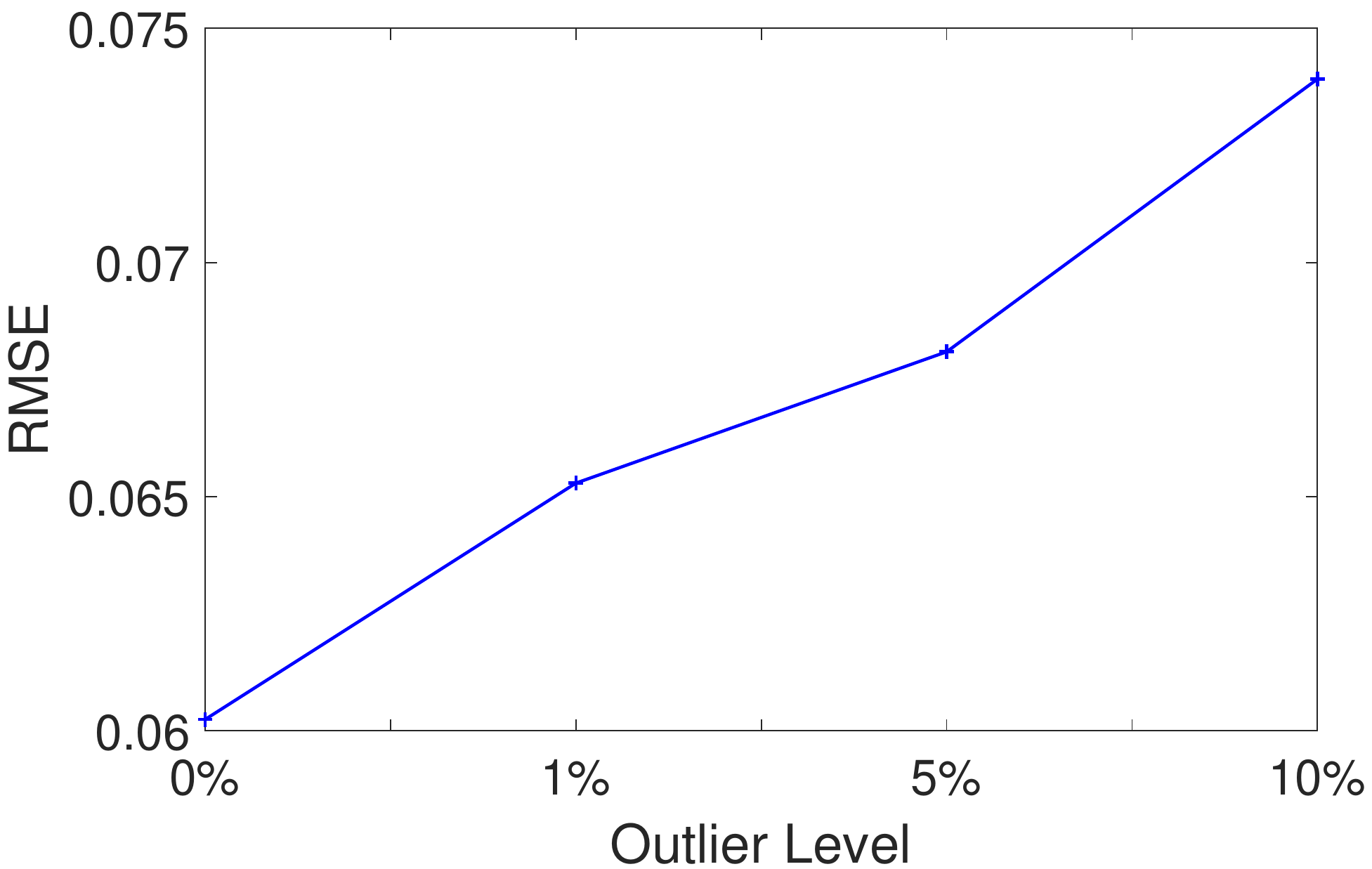}
  }
  \vspace{3pt}
  \vspace{3pt}
  \vspace{-10pt}
  \centering
  \caption{\normalsize {\color{black}PP-HFNN's performance} on an artificial dataset with different levels of noise (left) and outliers (right) }
  \label{fig:noise_analysis}
\end{figure*}
}
\subsection{Classification tasks}
As mentioned, we evaluated PP-HFNN’s {\color{black}classification} performance on the HRSS and EEG-DD datasets. Mean average precision (mAP) was the assessment metric, and we compared the results to SVM with different kernel functions. {\color{black}In addition, we also compared PP-HFNN’s with a standard non-hierarchical FNN, and an HFNN with a hierarchical structure but not implemented in a distributed scheme.} With the HRSS dataset, we randomly segmented the samples into five subsets, each of which was assigned to an agent. Each generated subset was then further split into six subsets according to the feature sources and subjected to five-fold cross-validation.

The features in the EEG-DD dataset relate to the different channels and bands of the signals and, thus, are heterogeneous. However, samples from the same trial are not independent and, therefore, the training data and the test data should not be from the same trial. Also, the EEG-DD dataset is complex with 11 subjects, hundreds of trials, and each trial separated into 17 segments. {\color{black}
We adopted the leave-one-out cross-validation method for each subject of the EEG-DD dataset. The mAP value reported is the averaged accuracy of all the subjects.} Likewise, with the HRSS dataset, we assigned each subject’s data to five agents and report the performance over all subjects {\color{black}in terms of} mAP.

\begin{table}[!ht]
  \centering
  \caption{Simulation results of the classification tasks}
  \resizebox{\columnwidth}{!}{
  \begin{tabular}{|c|c|c|c|c|c|c|} \hline
  \multirow{2}{*}{Dataset} & \multirow{2}{*}{Algorithm}  & \multirow{2}{*}{AO} & \multicolumn{4}{|c|}{Performance}     \\ \cline{4-7}
                            &             &      &Train(\%)            & Test(\%)              & t-test    & Time (s)   \\ \hline
  \multirow{3}{*}{HRSS}     & SVM(linear) & -    & 78.23/0.62          & 78.10/0.60            & 0.0001    & 2.83           \\ \cline{2-7}
                            & SVM(rbf)    & -    & 90.32/0.42          & 79.28/0.89            & 0.0467    & 3.19           \\ \cline{2-7}
                            & RF          & -    & 79.08/0.55          & 78.57/1.21            & 0.0159    & 4.17           \\ \cline{2-7}
                            & FNN         & -    & 79.86/0.54          & 76.11/0.88            & 0.0000    & 9.15           \\ \cline{2-7}
                            & HFNN        & 15   & \textbf{81.15/0.59} & \textbf{80.96/1.04}   & -    & 15             \\ \cline{2-7}
                            & PP-HFNN     & 15   & \textbf{80.26/0.57} & \textbf{80.30/0.37}   & -        & 18             \\ \hline
  \multirow{3}{*}{EEG-DD}   & SVM(linear) & -    & 83.80/2.51          & 76.20/2.41            &0.0000    & 0.24           \\ \cline{2-7}
                            & SVM(rbf)    & -    & 87.41/7.37          & 78.06/4.55            &0.0012    & 0.31           \\ \cline{2-7}
                            & RF          & -    & 87.55/3.26          & 81.03/2.20            &0.0098    & 1.21           \\ \cline{2-7}
                            & FNN         & -    & 87.68/2.04          & 77.44/2.03            &0.0000    & 12.21          \\ \cline{2-7}
                            & HFNN     & 20   & \textbf{87.48/4.34} & \textbf{84.94/5.42}   & -    & 16             \\ \cline{2-7}
                            & PP-HFNN     & 20   & \textbf{87.02/4.70} & \textbf{84.98/4.02}   & -       & 22            \\ \hline
  \end{tabular}}
  \label{cls_map}
\end{table}

The results for this classification task are provided in Table II. {\color{black} The number of iterations of the AO algorithm is given in the third column.} {\color{black}. The training and testing accuracy and corresponding standard variance values are provided in the fourth and fifth columns, respectively.} HFNN and PP-HFNN show a better mAP than the other algorithms, which proves that HFNNs can deal with heterogeneous data relatively well. The comparisons between centralized and distributed clustering show that HFNN performed better with a distributed architecture on the EEG-DD dataset, but slightly worse with the HRSS dataset. Overall, however, the results show that a low-level FNN with distributed clustering works better than using a central agent.
SVM is a powerful binary classification algorithm but, according to the results in Table II, PP-HFNN has the potential to be better.  As  Table  II shows, PP-HFNN returned a higher mAP than SVM on the EEG DD dataset, which has more heterogeneity than HRSS. From this, we conclude that PP- HFNN is a better option than SVM for highly heterogeneous datasets. {\color{black} In addition, we applied the well-known t-test to evaluate the statistical significance between PP-HFNN and the other methods.  The p-values of the t-tests are given in Column 6 of Table \ref{cls_map}. {\color{black} All p-values were less than 0.05, which indicates that the} performance improvements brought by PP-HFNN are statistically significant.}

\begin{figure*}[!ht]
  \centering
  \subfigure[\normalsize mAP of HFNN on Training data]{
  \includegraphics[width=0.8\columnwidth]{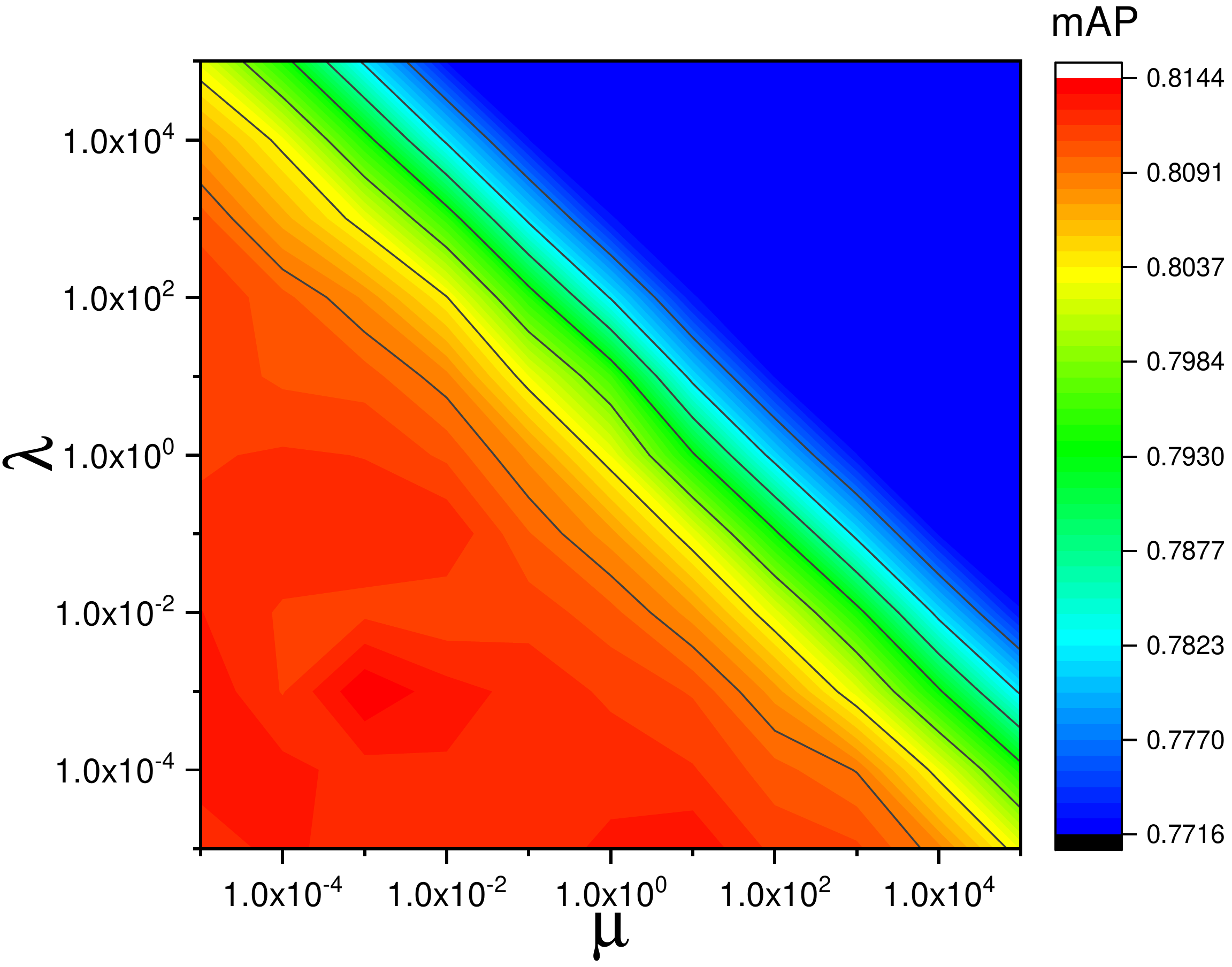}
  }
  \vspace{3pt}
  \hspace{3pt}
  \subfigure[\normalsize  mAP of HFNN on Test data]{
  \includegraphics[width=0.8\columnwidth]{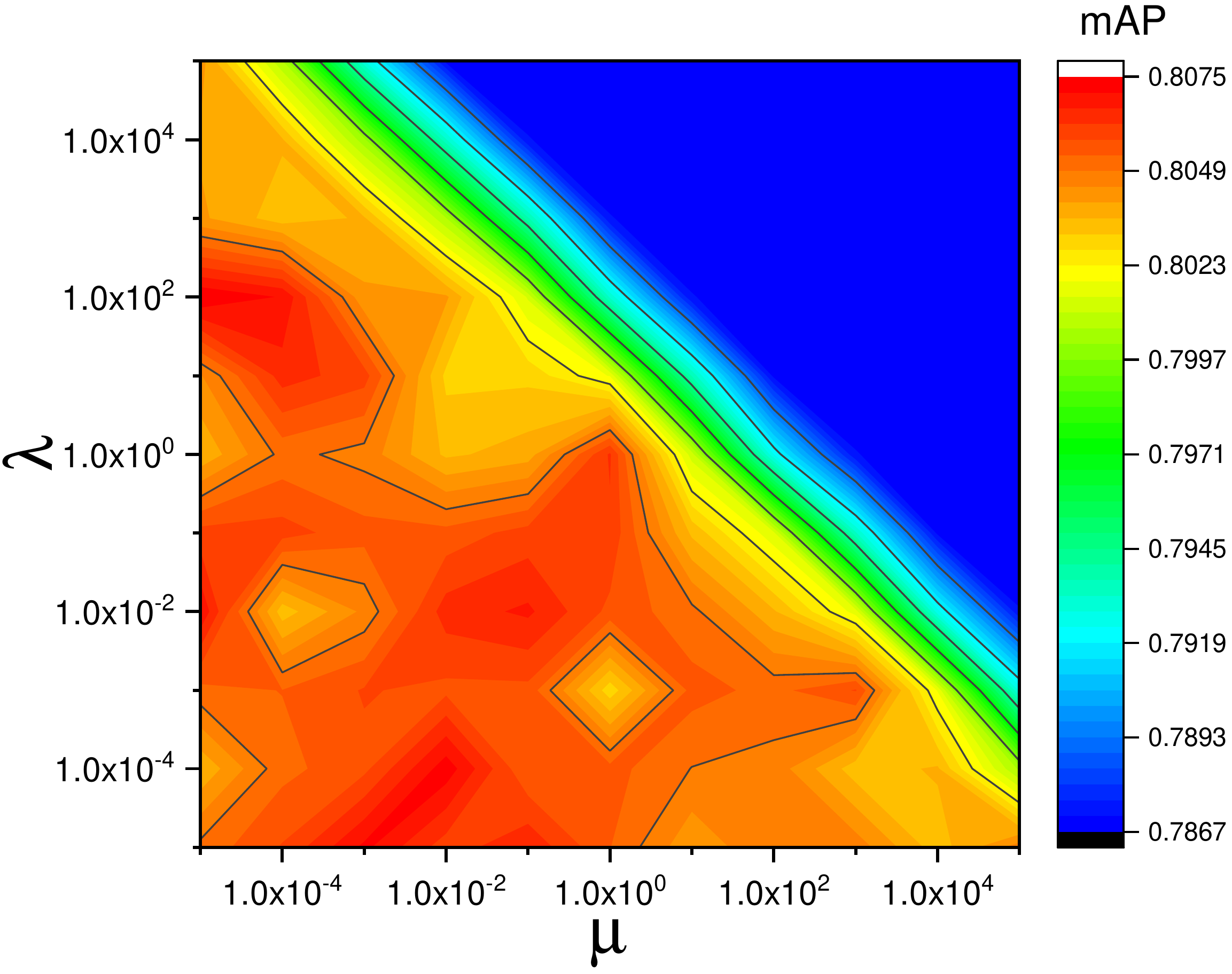}
  }
  \vspace{3pt}
  \hspace{3pt}
  \subfigure[\normalsize mAP of PP-HFNN on Training data]{
  \includegraphics[width=0.8\columnwidth]{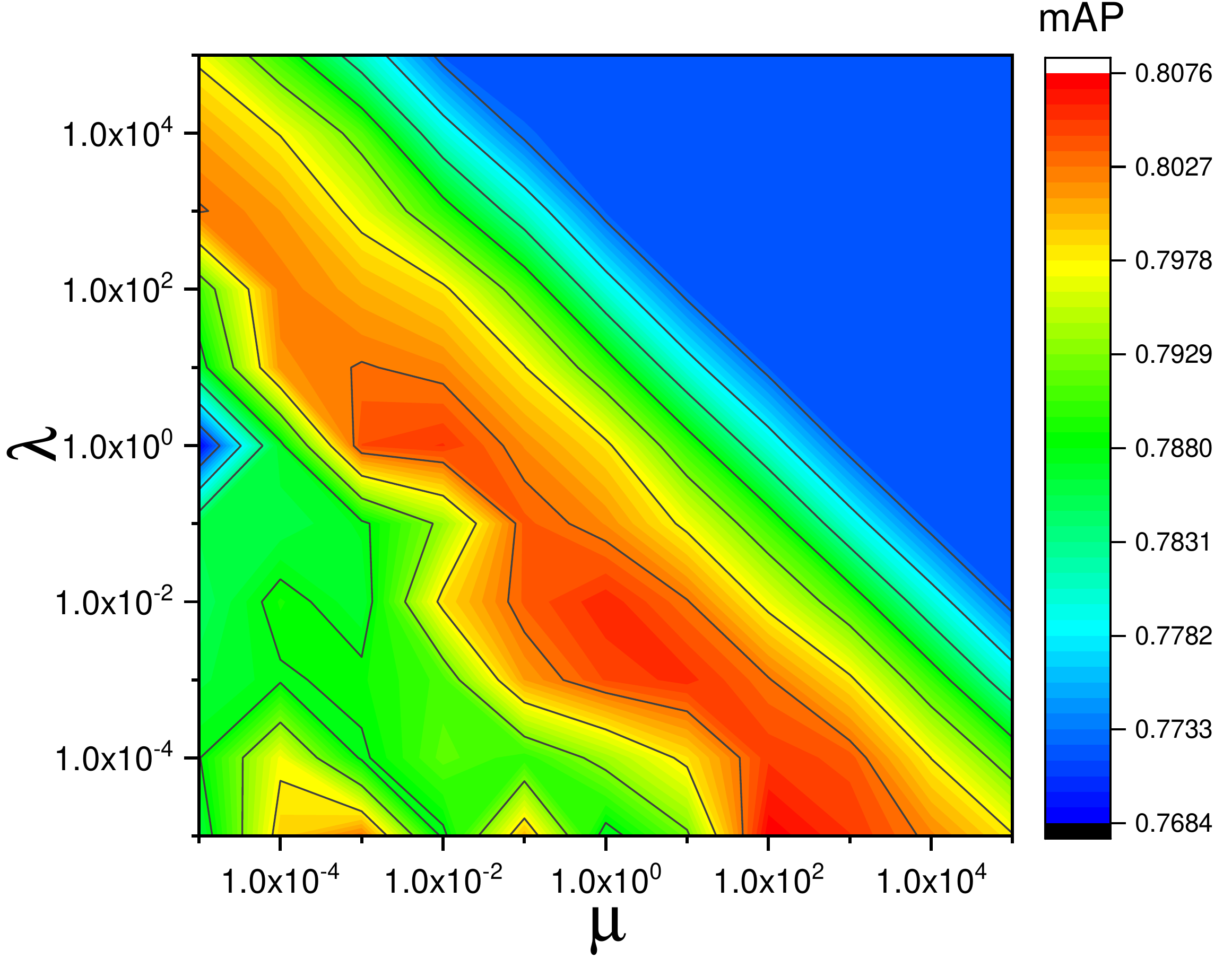}
  }
  \vspace{3pt}
  \vspace{3pt}
  \subfigure[\normalsize mAP of PP-HFNN on Test data]{
  \includegraphics[width=0.8\columnwidth]{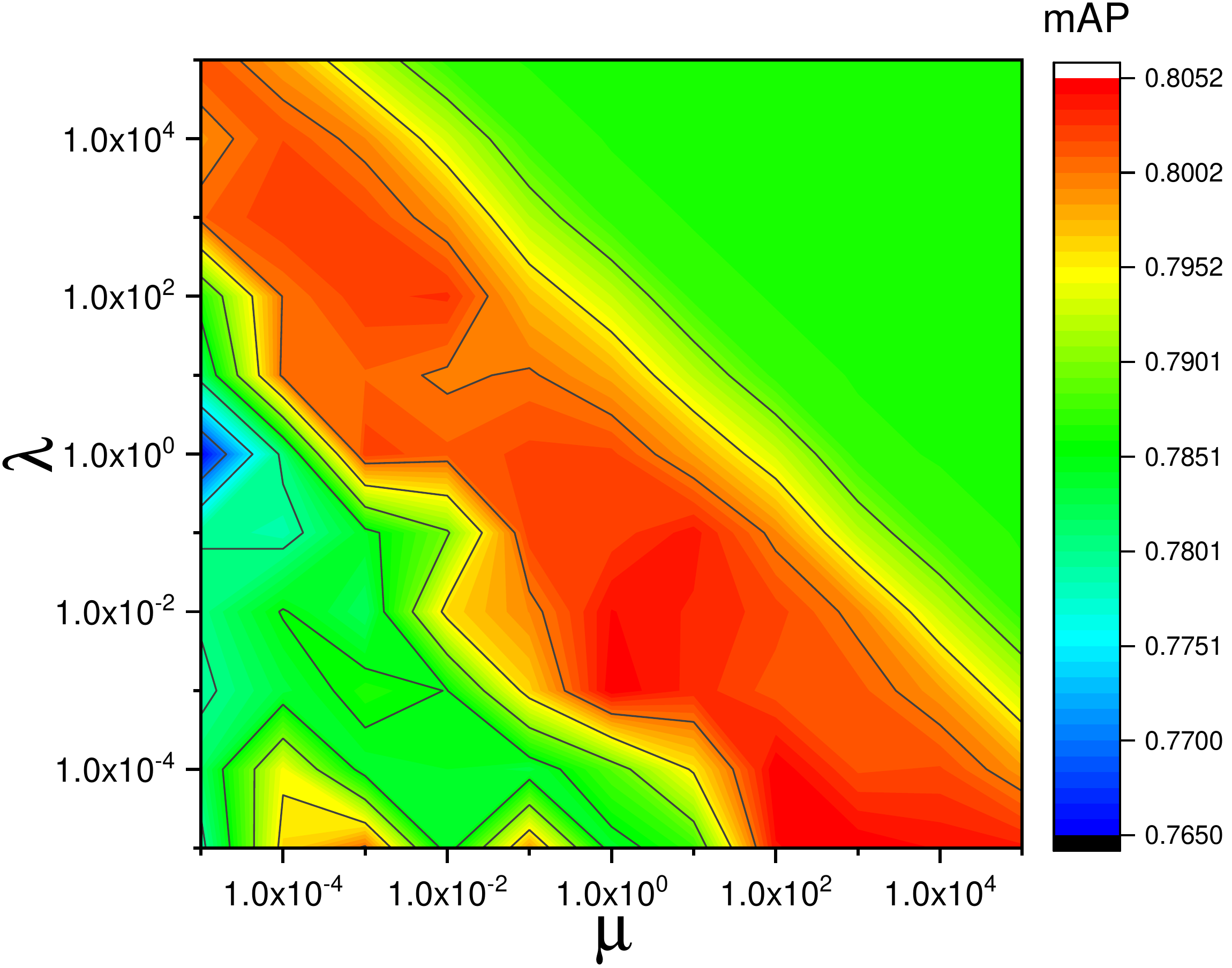}
  }
  \vspace{3pt}
  \vspace{3pt}
  \vspace{-10pt}
  \centering
  \caption{\normalsize Performance of different parameter settings on the HRSS dataset, with each low-level FNN has 4 rules. Results of using HFNN on the (1) training data and (2) test data, and the results of using PP-HFNN on the (1) training data and (2) test data}
  \label{fig:para_analysis}
\end{figure*}

\subsection{Regression tasks}
To assess PP-HFNN’s performance with regression tasks, we compared it to linear regression, Lasso regression, Ridge regression, polynomial regression and support vector regression (SVR) with the EC dataset and five-fold cross-validation. The degree of polynomial regression ranged from 2 to 5. The evaluation metric was mean square error (MSE), calculated by

\begin{equation}\label{RMSE}
  \mbox{MSE} = \frac{1}{N\hat{\sigma}_Y}\sum_{i=1}^{N}(\hat{Y}_i-Y_i)^2,
\end{equation}
The results are listed in Table \ref{tab:reg_loss}.

\begin{table}[!ht]
  \centering
  \caption{Simulation results of the regression tasks}
  \resizebox{\columnwidth}{!}{
  \begin{tabular}{|c|c|c|c|c|c|c|} \hline
  \multirow{2}{*}{Output} & \multirow{2}{*}{Algorithm}  & \multirow{2}{*}{AO} & \multicolumn{4}{|c|}{Performance}     \\ \cline{4-7}
                            &           &      &Train                      & Test                      & t-test      & Time(s)  \\ \hline
  \multirow{3}{*}{CO}       & Linear    & -    & 0.1145/0.0057             & 0.1139/0.0077             & 0.0001  & 11      \\ \cline{2-7}
                            & Lasso     & -    & 0.1559/0.0050             & 0.2025/0.0139             & 0.0000  & 12      \\ \cline{2-7}
                            & Ridge     & -    & 0.1176/0.0049             & 0.1199/0.0174             & 0.0016  & 49      \\ \cline{2-7}
                            & Polynomial& -    & 0.1034/0.0067             & 0.1003/0.0075             & 0.0026 & 2937    \\ \cline{2-7}
                            & RF        & -    & 0.0999/0.0043             & 0.1031/0.0085             & 0.0018  & 1851    \\ \cline{2-7}
                            & HFNN      & 50   & \textbf{0.0827/0.0014}    & \textbf{0.0861/0.0067}    & - & 3431    \\ \cline{2-7}
                            & PP-HFNN   & 50   & \textbf{0.0770/0.0058}    & \textbf{0.0806/0.0069}    & -      & 1242    \\ \hline
  \multirow{3}{*}{Ethylene} & Linear    & -    & 0.1139/0.0061             & 0.1274/0.0117             & 0.0000  & 11      \\ \cline{2-7}
                            & Lasso     & -    & 0.1673/0.0063             & 0.1634/0.0143             & 0.0000  & 12      \\ \cline{2-7}
                            & Ridge     & -    & 0.1061/0.0062             & 0.0974/0.0124             & 0.0152  & 49      \\ \cline{2-7}
                            & Polynomial& -    & 0.1076/0.0057             & 0.1035/0.0191             & 0.0215  & 2937    \\ \cline{2-7}
                            & RF        & -    & 0.0999/0.0050             & 0.0979/0.0204             & 0.0290  & 1851    \\ \cline{2-7}
                            & HFNN      & 50   & \textbf{0.0724/0.0036}    & \textbf{0.0782/0.0058}    & -  & 3431    \\ \cline{2-7}
                            & PP-HFNN   & 50   & \textbf{0.0738/0.0041}    & \textbf{0.0768/0.0084}& -       & 1242    \\ \hline
  \end{tabular}}
  \label{tab:reg_loss}
\end{table}

As shown, HFNN performed better than all the other regression methods, which indicates that a hierarchical structure is more effective for processing heterogeneous data than a non-hierarchical structure. Also, we observed that using a distributed method took almost half the time of the centralized method, which is a significant improvement in time and efficiency. Notably, SVR could not cope with high volumes of data and failed to produce any output. {\color{black} We conducted a t-test for this experiment, too. Again, all p-values were than 0.05 (see column 6 Table \ref{tab:reg_loss}), confirming statistical significance of the performance improvements brought by PP-HFNN.}

\subsection{Results analysis}
\begin{figure*}[!h]
  \centering
  \subfigure[\normalsize MSE value of HRSS]{
  \includegraphics[width=0.30\textwidth]{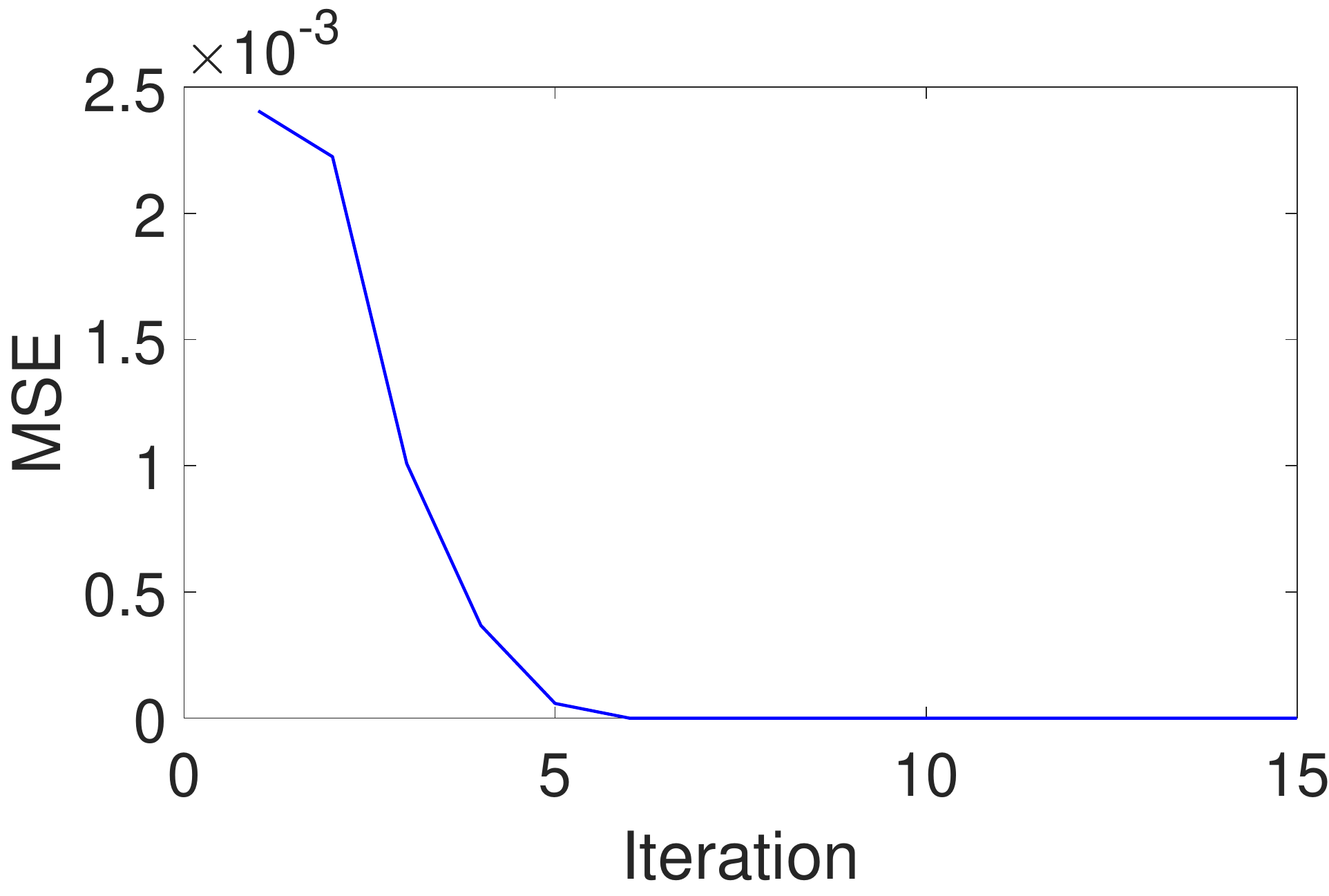}
  }
  \vspace{3pt}
  \hspace{3pt}
  \subfigure[\normalsize MSE value of EEG-DD]{
  \includegraphics[width=0.30\textwidth]{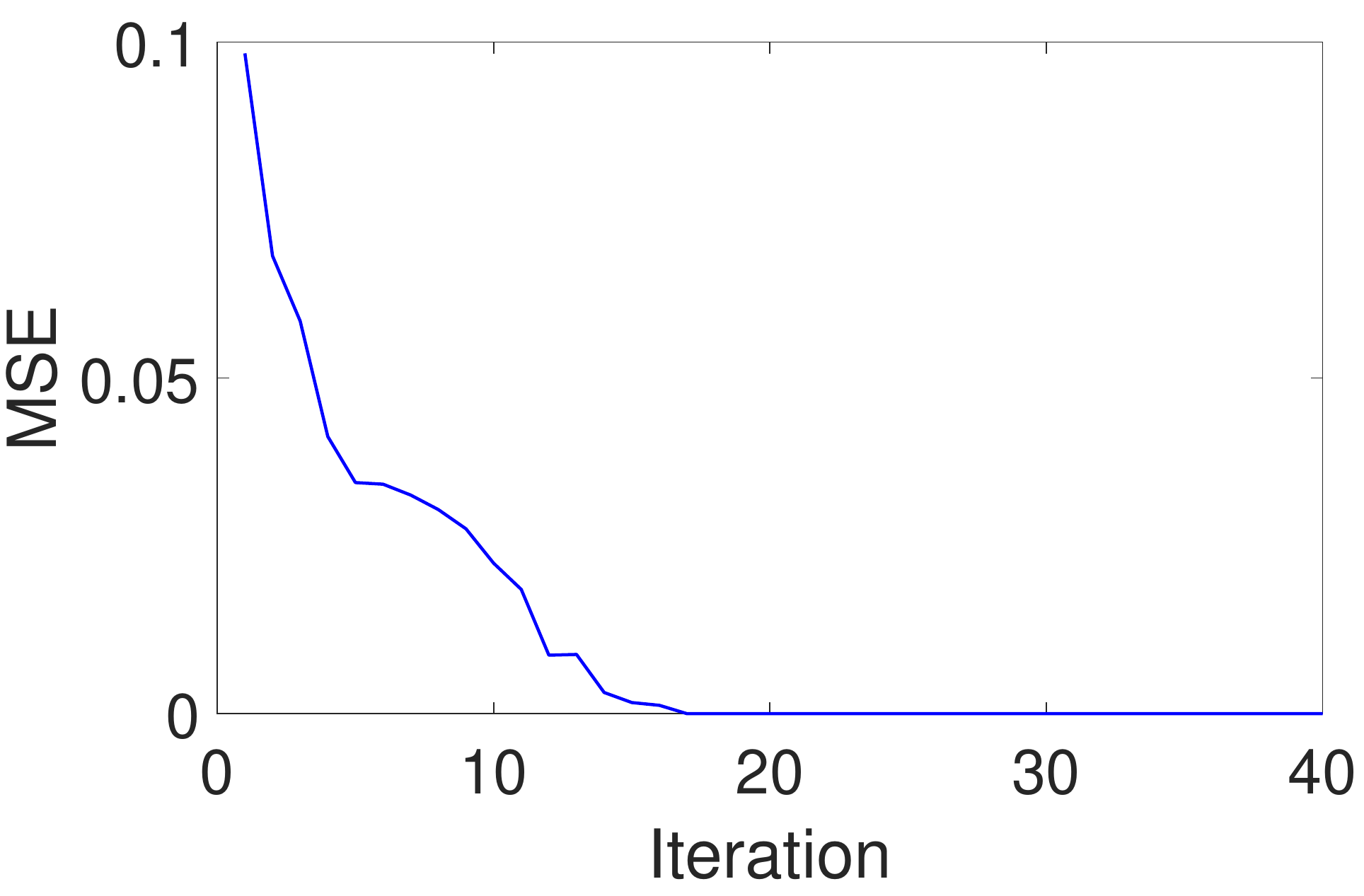}
  }
  \vspace{3pt}
  \vspace{3pt}
  \subfigure[\normalsize  MSE value of Ethylene-CO]{
  \includegraphics[width=0.30\textwidth]{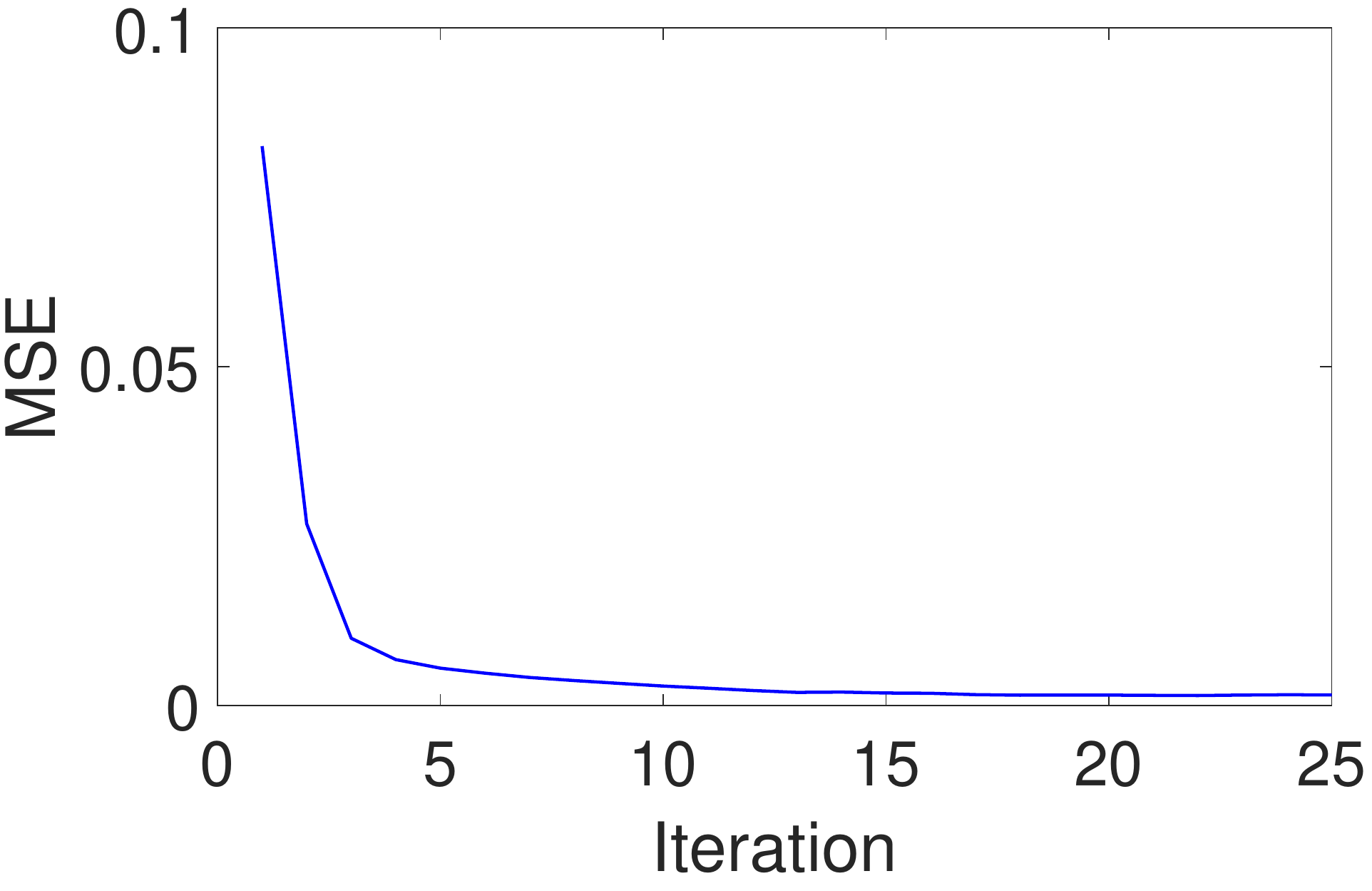}
  }
  \vspace{3pt}
  \vspace{3pt}
  \caption{\normalsize MSE value of distributed K-means using the ADMM.}
  \label{fig:dkmm_loss}
\end{figure*}

\begin{figure}[!t]
  \centering
  \includegraphics[width=1\columnwidth]{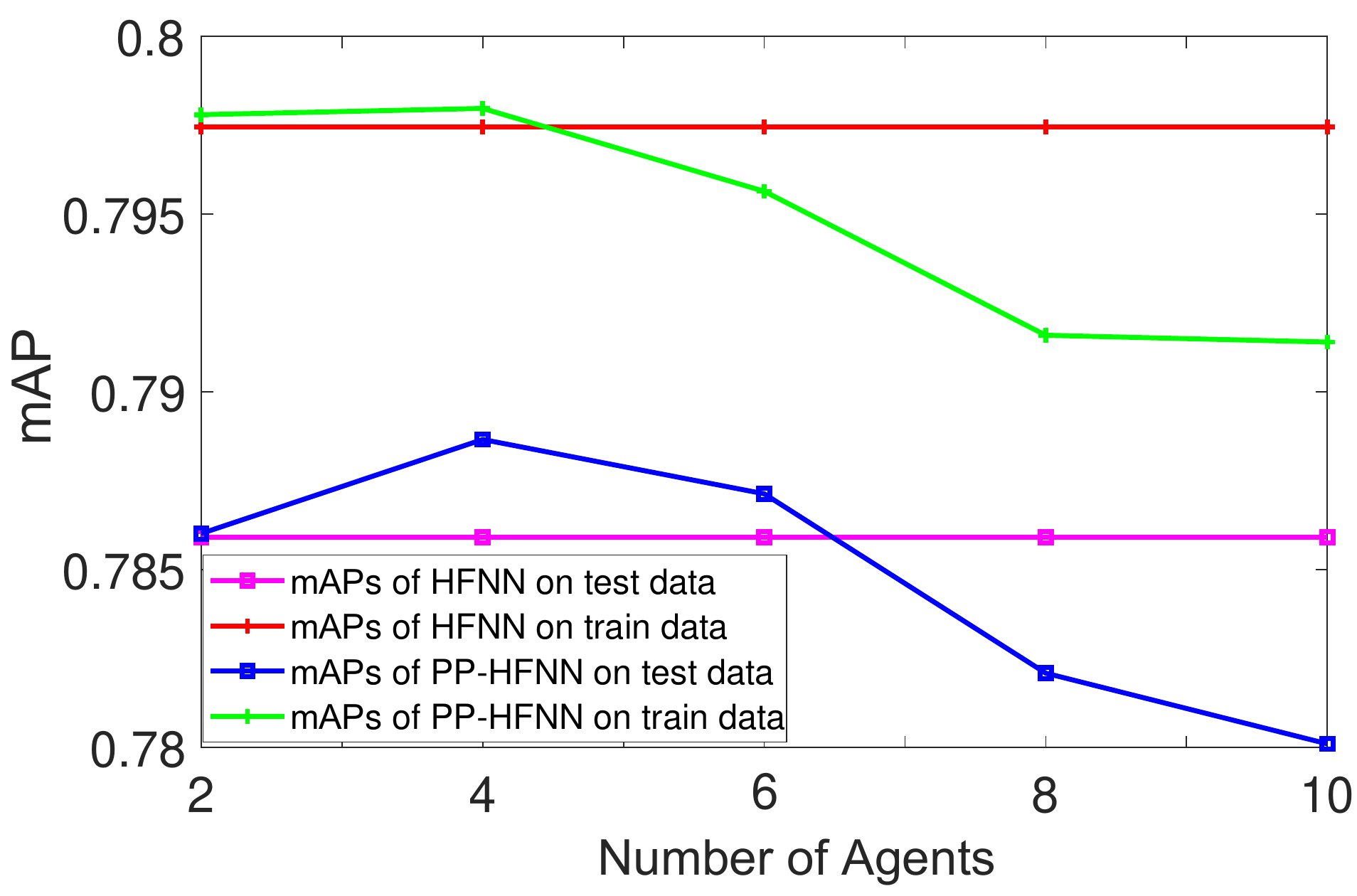}
  \vspace{-10pt}
  \caption{The performance of PP-HFNN with different number of agents.}
  \label{fig:agent_hrss}
\end{figure}

{\color{black} It can be seen from Tables \ref{cls_map} and  \ref{tab:reg_loss} that PP-HFNN reached higher accuracy with both the classification and regression tasks than the other tested methods. We believe there are three reasons{\color{black}for this}: 1) its hierarchical structure, which extracts data features from heterogeneous big data and also reduces the number of fuzzy rules; 2) the distributed K-means algorithm, which is efficient with large-scale data sets and also protects data privacy; {\color{black}and} 3) the alternating optimization algorithm, which converges very {\color{black}quickly} and does not suffer from the vanishing gradient problem associated with gradient-based learning methods.

We also investigated the trade-off between privacy-preservation and performance by increasing the number of agents from 2 to 10 for the HRSS dataset. The results, presented in Fig. \ref{fig:agent_hrss}, show that the mAP value for PP-HFNN initially increases and then drops as the number of agents increases. In addition, PP-HFNN's performance on the testing data was even better than its centralized counterpart HFNN with less than six agents. Note that the difference between PP-HFNN and HFNN is that PP-HFNN employs a distributed K-means algorithm, while HFNN uses a centralized K-means algorithm to locate the parameters in the antecedent layer. Here, the distributed K-means algorithm can be regarded as a distributed representation learning process, which, in some cases, can outperform its centralized counterpart. However, since the number of training samples allocated to each agent decreases as the number of agents increases, in this case, the accuracy of PP-HFNN dropped with greater data privacy protection.}

To analyze the influence of the parameter settings, {\color{black} we plotted the mAPs for the HRSS datasets using four rules on each low-level FNN over the parameter space.} As shown in Fig. \ref{fig:para_analysis}, the optimal parameters for PP-HFNN are located in a diagonal area.  Additionally, PP-HFNN’s mAP stayed above 77\% no matter which parameter was used. This is because the low-level FNNs {\color{black}are able to} extract informative feature representations, which means the AO method is not sensitive to the parameters. Fig. \ref{fig:dkmm_loss} shows the MSE value curves for the distributed K-means algorithm with all three datasets to analyze convergence. All converged within 20 iterations.

\section{Conclusion}
The last few decades have witnessed a big data boom in many areas, such as social networks, the Internet of Things, commerce, astronomy, biology, medicine, etc. However, heterogeneous big data poses a range of challenges to machine learning given its enormous scale and dimensionality and its inherent uncertainty. In this paper, we proposed a privacy-preserving HFNN to address some of the challenges with heterogeneous big data, and especially those surrounding privacy concerns. We presented a two-stage optimization algorithm to train the HFNN. The parameters at low levels of the hierarchy are trained by {\color{black}a} distributed K-means algorithm, which does not reveal local data among agents and, thus, is privacy-preserving. Additionally, the distributed K-means algorithm has very good performance, especially with heterogeneous big data {\color{black}at massive scales}. Coordination at high levels of the hierarchy is conducted with the AO method, which converges very {\color{black}quickly}. The entire training procedure is scalable and does not suffer from slow training speeds or gradient vanishing problems, unlike the methods based on back-propagation. Comprehensive simulations on both regression and classification tasks show that this PP-HFNN outperforms all the tested baseline methods in terms of accuracy and, further, the difference in training time widens as the size of the dataset  grows. {\color{black} We are currently exploring a semi-supervised learning method based on the PP-HFNN framework to address scenarios where the training data consists of both labeled and unlabeled samples.}
\bibliographystyle{ieeetr}
\bibliography{PP-HFNN}

\begin{IEEEbiography}[{\includegraphics[width=1in,height=1.25in,clip,keepaspectratio]{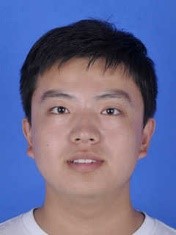}}]
{Leijie Zhang} received the B.S. degree from Hebei Normal University, China, in 2015 and the M.S. degree from Hangzhou Dianzi University,  China, in 2019. He is currently pursuing the Ph.D. degree in computer science with the University of Technology Sydney, Ultimo, NSW, Australia. His current research interests include fuzzy neural networks and reinforcement learning.
\end{IEEEbiography}

\vspace{-1cm}

\begin{IEEEbiography}[{\includegraphics[width=1in,height=1.25in,clip,keepaspectratio]{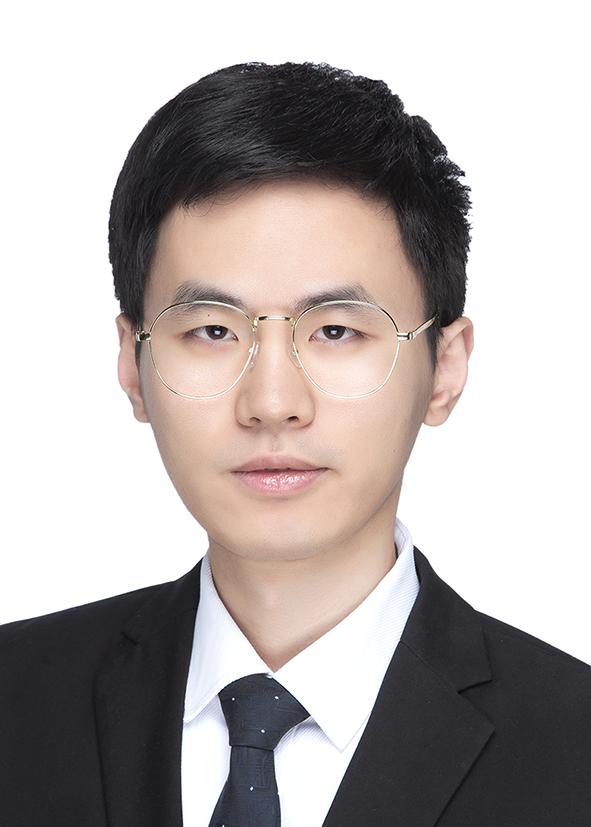}}]
{Ye Shi} (M’19) received the B.S. degree in Statistics from Northwestern Polytechnical University, China in 2013 and the Ph.D. degree in electrical engineering at University of Technology Sydney, Australia in 2018. He has been a postdoctoral fellow with the school of computer science at University of Technology Sydney since 2019. His research interests include mixed-integer nonlinear programming, spectral optimization and fuzzy neural networks. He received the Best Paper Award at the 6th IEEE International Conference on Control Systems, Computing and Engineering in 2016.
\end{IEEEbiography}

\vspace{-1cm}

\begin{IEEEbiography}[{\includegraphics*[width=1in, height=1.25in, clip, keepaspectratio]{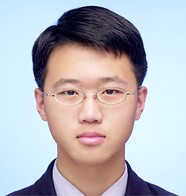}}]
{Yu-Cheng Chang} received the B.S. degree in vehicle engineering from the National Taipei University of Technology, Taipei, Taiwan, in 2008, the M.S. degree with a specialization in system and control from the Department of Electrical Engineering, National Chung-Hsing University (NCHU), Taichung, Taiwan, in 2010. From 2014 to 2016, He had been a Research Assistant with the Department of Electrical Engineering, NCHU. He is perusing his Ph.D. degrees in computer science at University of Technology Sydney since 2016. His current research interests include fuzzy systems, human-machine autonomous, and brain state analysis.
\end{IEEEbiography}

\vspace{-1cm}

\begin{IEEEbiography}[{\includegraphics[width=1in,height=1.25in,clip,keepaspectratio]{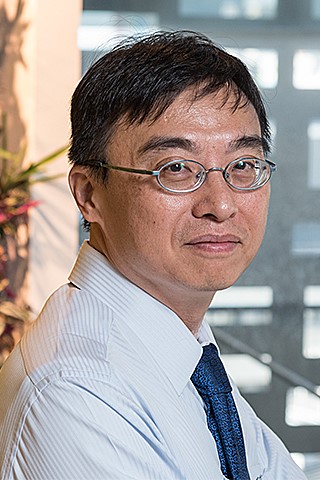}}]
{Chin-Teng Lin} (F’05) received the B.S. degree from National Chiao-Tung University (NCTU), Taiwan in 1986, and the Master and Ph.D. degree in electrical engineering from Purdue University, USA in 1989 and 1992, respectively. He is currently the Distinguished Professor of Faculty of Engineering and Information Technology, and Co-Director of Center for Artificial Intelligence, University of Technology Sydney, Australia. He is also invited as Honorary Chair Professor of Electrical and Computer Engineering, NCTU,  and Honorary Professorship of University of Nottingham. Dr. Lin was elevated to be an IEEE Fellow for his contributions to biologically inspired information systems in 2005, and was elevated International Fuzzy Systems Association (IFSA) Fellow in 2012. Dr. Lin received the IEEE Fuzzy Systems Pioneer Awards in 2017. He served as the Editor-in-chief of IEEE Transactions on Fuzzy Systems from 2011 to 2016. He also served on the Board of Governors at IEEE Circuits and Systems (CAS) Society in 2005-2008, IEEE Systems, Man, Cybernetics (SMC) Society in 2003-2005, IEEE Computational Intelligence Society in 2008-2010, and Chair of IEEE Taipei Section in 2009-2010. Dr. Lin was the Distinguished Lecturer of IEEE CAS Society from 2003 to 2005 and CIS Society from 2015-2017. He serves as the Chair of IEEE CIS Distinguished Lecturer Program Committee in 2018~2019. He served as the Deputy Editor-in-Chief of IEEE Transactions on Circuits and Systems-II in 2006-2008. Dr. Lin was the Program Chair of IEEE International Conference on Systems, Man, and Cybernetics in 2005 and General Chair of 2011 IEEE International Conference on Fuzzy Systems. Dr. Lin is the coauthor of Neural Fuzzy Systems (Prentice-Hall), and the author of Neural Fuzzy Control Systems with Structure and Parameter Learning (World Scientific). He has published more than 330 journal papers (Total Citation: 22,913, H-index: 68, i10-index: 290) in the areas of neural networks, fuzzy systems, brain computer interface, multimedia information processing, and cognitive neuro-engineering, including about 125 IEEE journal papers.
\end{IEEEbiography}

\end{document}